\PassOptionsToPackage{dvipsnames}{xcolor}
\documentclass{article}
\usepackage{amsmath, amssymb, amsthm, amsfonts}
\usepackage{lmodern}
\usepackage{arxiv}
\usepackage{physics}
\usepackage[T1]{fontenc}
\usepackage{graphicx}
\usepackage{wrapfig} %
\usepackage{subcaption}
\usepackage{authblk}
\usepackage{cancel}
\usepackage{tcolorbox}

\usepackage{booktabs}
\usepackage{multirow}
\usepackage[table]{xcolor}

\definecolor{proxblue}{HTML}{75ADD4}
\definecolor{quadorange}{HTML}{FF9F50}
\definecolor{batchshade}{HTML}{F5F5F5}

\newcommand{\Prox}{\textcolor{proxblue}{\textsc{Prox}}}
\newcommand{\Quad}{\textcolor{quadorange}{\textsc{Quad}}}

\usepackage{parskip}
\newcommand\blfootnote[1]{%
  \begingroup
  \renewcommand\thefootnote{}%
  \footnotetext{#1}%
  \endgroup
}

\usepackage{amsmath,amsfonts,bm}

\def\eqref#1{equation~\ref{#1}}

\def\1{\mathbf{1}}

\DeclareMathAlphabet{\mathsfit}{\encodingdefault}{\sfdefault}{m}{sl}
\SetMathAlphabet{\mathsfit}{bold}{\encodingdefault}{\sfdefault}{bx}{n}

\newcommand{\E}{\mathbb{E}}

\newcommand{\R}{\mathbb{R}}

\newcommand{\Cov}{\mathrm{Cov}}

\newtheorem{proposition}{Proposition}

\newtheorem{definition}{Definition}

\newcommand{\lprox}{\mathcal{L}_{\mathrm{prox}}}
\newcommand{\lquad}{\mathcal{L}_{\mathrm{quad}}}

\newcommand{\why}[1]{\hspace{2em}\textcolor{gray}{\text{#1}}}
\newcommand{\tok}[1]{%
  \begingroup
  \setlength{\fboxsep}{1.5pt}%
  \fcolorbox{gray!35}{gray!10}{\texttt{\strut\,#1\,}}%
  \endgroup
}

\usepackage[utf8]{inputenc} %
\usepackage{enumitem}
\usepackage[T1]{fontenc}    %
\usepackage{hyperref}       %
\usepackage{url}            %
\usepackage{booktabs}       %
\usepackage{amsfonts}       %
\usepackage{nicefrac}       %
\usepackage{microtype}      %
\usepackage{xcolor}         %
\usepackage{cleveref}
\crefname{appendix}{appendix}{appendices}
\Crefname{appendix}{Appendix}{Appendices}
\usepackage{algorithm}
\usepackage{algpseudocode}

\usepackage{titlesec}
\titlespacing*{\paragraph}{0pt}{3.25ex plus 1ex minus .2ex}{1em}

\hypersetup{colorlinks=true,allcolors=NavyBlue}

\title{A Defense of the Quadratic Model}
\author[1]{Alexandru Meterez}
\author[1]{Pranav Ajit Nair}
\author[1,$\dagger$]{Depen Morwani}
\author[1]{Cengiz Pehlevan}
\author[1]{\\Sham Kakade}
\author[1,2]{Alex Damian}

\affil[1]{Kempner Institute at Harvard University}
\affil[2]{MIT}
\begin{document}

\maketitle

\begin{abstract}
Due to the complexity of neural network loss landscapes, optimization theory is forced to rely on idealized models, and there is generally a tradeoff between how theoretically tractable the model is, and how accurately it describes the true optimization dynamics. In this work, we stress test the simplest possible model of optimization -- the quadratic model -- and show that it can be surprisingly predictive in an LLM setting with 150M parameters and 3B training tokens. Specifically, we show that Taylor expanding the model and the loss function at intermediate checkpoints through training can accurately predict the optimization dynamics over windows that can last up to 10\% of training. Having established this agreement, we then turn to analyzing the structure of these local quadratic optimization problems through two lenses: the Hessian spectrum and local stability. Using Lanczos quadrature with extremely deep probes, we are able to estimate the Hessian spectrum deep into the tail, and we find a surprising amount of structure in both the eigenvalues and eigenvectors, which depends on the batch size, preconditioner, and training time. We also empirically test local linear stability at intermediate checkpoints and compare it to theoretical predictions to demonstrate that optimization in LLMs typically occurs at a stochastic edge of stability, whose nature is also determined by batch size. Our results indicate the quadratic model may be a theoretically tractable proxy for pretraining optimization dynamics.

\end{abstract}

\section{Introduction}
\label{sec:intro}
\blfootnote{\hspace{-6mm} $\dagger$: Work done while at Harvard.}
Large Language Model (LLM) optimization has quickly advanced through a feedback loop between fast experimentation and theoretical intuition. For pretraining, many design choices such as learning rate scheduling, batch-size scaling, and optimizer are, at least informally, motivated by traditional optimization theory~\citep{polyak1964some,zhang2024does,polyak1992acceleration,meterez2025seesaw,kidambi2018insufficiency}. Such models generally act as a useful guide for reasoning about training dynamics and scaling in a tractable way. However, it is unclear which of these models are actually descriptive of the dynamics of large-scale pretraining. 

There are several failure modes that give rise to this gap between theoretical proxies and empirics. On one hand, models can be overly idealized: they can capture a mathematical phenomenon, but under assumptions that are difficult to justify for modern neural networks. However, if the model does not incorporate enough structure about the specific optimization problem (e.g. worst case bounds over the set of all smooth convex problems), it can lead to predictions that are overly pessimistic and do not lead to actionable insights. Finally, if a model is not overly idealized and incorporates problem-specific structure, it is often theoretically intractable.%

Arguably the simplest setting in which optimization theory can give sharp problem-specific predictions is the quadratic model. The advantage of this setting is that the training dynamics can be understood mode-by-mode across the Hessian spectrum, which enables instance-dependent analyses that are tight for a given Hessian spectrum. This makes it a useful sandbox for studying stability, learning rate scheduling, batch size scaling, acceleration, and preconditioning, and developing principled algorithms that transfer to neural network optimization~\citep{meterez2025seesaw,meterez2026anytime,zou2021benign,wu2018sgd,kidambi2018insufficiency,jain2018parallelizing,morwani2026compute,zhang2024does,zhang2019algorithmic,paquette2021dynamics,zhang2024optimality,wu2025risk,ferbach2026dimension,zhang2026scaling,varre2021last,varre2022accelerated,bordelon2021learning}.

These works largely use the quadratic model as a useful theoretically tractable sandbox for testing algorithms and ideas, and take a leap of faith when applying them to neural networks. In this work, we aim to tighten this connection and stress test the quadratic model by asking:
\begin{center}
    \emph{To what extent does the quadratic model actually capture the training dynamics of LLMs?}
\end{center}

In \Cref{sec:quadratics_are_predictive}, we test this by training a 150M transformer on 3B tokens from FineWeb and save intermediate checkpoints every 10\% of training. We then Taylor expand the model and the loss around each checkpoint and train on this proxy loss for 10\% of the total tokens, and compare this to the original optimization trajectory. We do this for two proxies. The first, \texttt{prox}, linearizes the model but maintains the cross-entropy objective, which is equivalent to multi-class logistic regression under the NTK~\citep{jacot2018neural,lee2019wide,chizat2019lazy}. The second, \texttt{quad}, Taylor expands this objective to second order so that the loss function is a convex quadratic. Our results indicate that these approximations can hold for up to 10\% of training, and the quality of the approximation is significantly better near the end of training than at the start.

Motivated by the accuracy of these Taylor expansions, we then turn to describing the structure of these local quadratic problems. In \Cref{sec:hessian_llms}, we estimate the spectra of the Hessian and the Gauss-Newton matrix along the training trajectory, both with and without the Adam preconditioner. To do so, we use extremely deep Lanczos quadrature probes with rigorous error bands and we are able to resolve the spectrum over six orders of magnitude. We find that the spectrum often splits into a head, whose size is determined by the vocab size and is dominated by the unembedding layer, and a power-law tail that is \emph{universal}: identical across batch sizes and unaffected by the Adam preconditioner. Based on these curves, we estimate the source and capacity exponents and find a capacity exponent of $\alpha \approx 1$ in the resolvable tail, and a source exponent of $\beta < 1$, which contradicts the summability assumption common in literature.

Finally, in \Cref{sec:edge_of_stability}, we analyze training stability by estimating how close each saved checkpoint is to the \emph{edge of stability}~\citep{defossez2015averaged, jain2018parallelizing, ma2018power, wu2018sgd, ma2021linear, wu2022alignment, velikanov2022view, mulayoff2024exact, cohen2021gradient, jastrzkebski2018relation, jastrzebski2020break, cohen2022adaptive, lewkowycz2020large, ahn2022understanding, arora2022understanding, damian2022self, cohen2024understanding}. This is a regime in which the linearized dynamics (preconditioned SGD) are incredibly sensitive to the learning rate and batch size, and any increase in the learning rate or decrease in the batch size could cause training to diverge for the linearized model. We empirically test this by running preconditioned SGD from various checkpoints with a grid of learning rate and batch size multipliers to test which hyperparameters would cause SGD to blow up, and we compare these results to the theoretical predictions made by linear stability analysis. We observe that at small to medium batch sizes, training occurs in a noise dominated stochastic edge of stability regime, while at large batch sizes, training operates at the deterministic edge of stability as described in \cite{cohen2021gradient}. We therefore conclude that the majority of LLM training occurs at the edge of stability, regardless of batch size.

\begin{figure}[!htp]
    \centering
    \includegraphics[width=1.0\linewidth]{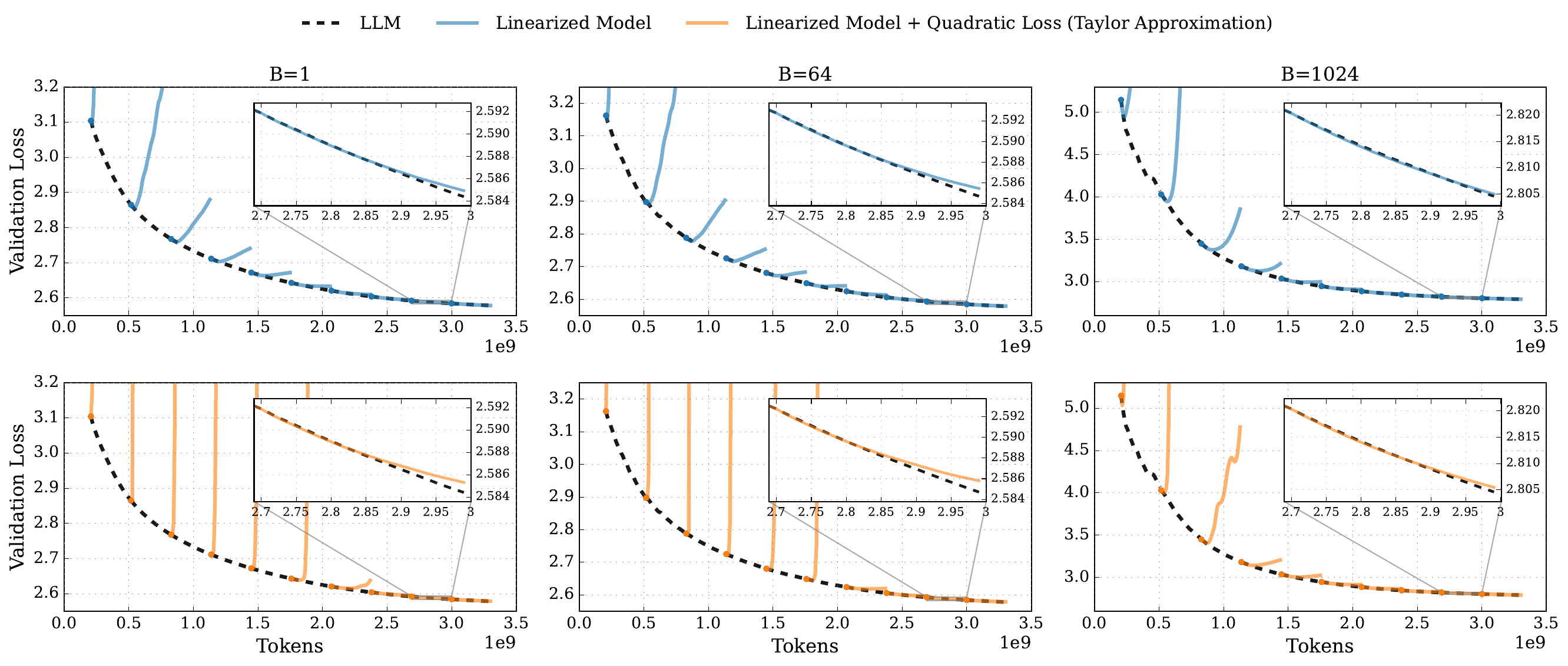}
    \caption{Validation loss curves for 150M models (black) trained with cosine for $1 \times$ Chinchilla tokens, across batch sizes, at optimal learning rates. (Top) We show agreement between the LLM loss curve when compared to training the linearized model around checkpoints recorded every $10 \%$ of training (blue). (Bottom) We further quadratic Taylor expand the loss and train the quadratic model (orange). The losses are still in agreement for late training times, but diverge at early times, as opposed to the case where we only linearize the model. Note that the orange and blue curves are trained on the expansions, but we evaluate them on the true model loss. We measure how long the approximation holds for from each checkpoint in~\Cref{tab:linearization-accuracy}. We provide further details in Section~\ref{sec:quadratics_are_predictive}, and we also show a similar plot for a constant learning rate with EMA in  Figure~\ref{fig:constant_linearization_loss_plots} (Appendix~\ref{app:constant_learning_rate}).}
    \label{fig:cosine_linearization_loss_plots}
\end{figure}

The full code for our experiments, along with all of the experimental data, is available at \url{https://github.com/alexandrumeterez/QuadraticModel}.

\section{Related Work}
\label{sec:related_work}
\paragraph{Linearized Dynamics.} The Neural Tangent Kernel
(NTK)~\citep{jacot2018neural,lee2019wide} approximates a network through its first-order Taylor expansion in parameter space, an approximation which is accurate whenever the parameter displacement is relatively small. This regime is also known as "lazy" training~\citep{chizat2019lazy}. However, in practice, the parameters and the NTK evolve rapidly as features form~\citep{fort2020deep,atanasov2021neural,long2021properties,vyas2022limitations, ortiz2021can}, rendering the NTK dynamics inaccurate for practical learning rates and widths. Several other works have shown that using the "after kernel", meaning the linearized model at the end of training, is a more faithful approximation for interpretability and fine-tuning~\citep{park2023trak, long2021properties,malladi2023kernel}. A complementary line of work studies higher-order Taylor expansions~\citep{bai2019beyond, bai2020taylorized} as a better approximation for the true dynamics. Relative to this literature, we linearize locally throughout the trajectory rather than at initialization, and we measure the agreement window as a fraction of the total training budget.

\paragraph{Hessian Spectra.} Neural network Hessian spectra have been previously studied in order to understand the optimization loss landscape. While for small models it is possible to compute Hessian eigenvalues exactly~\citep{sagun2016eigenvalues, sagun2017empirical}, either by calculating the Hessian in closed form or through a relatively small number of power iterations, this quickly breaks down when parameter counts are in the millions. A more refined algorithm, which is also the basis of \Cref{sec:spectrum}, is the Stochastic Lanczos Quadrature (SLQ) algorithm ~\citep{lanczos1950iteration, golub2009matrices}. Since its introduction, numerous works have studied this algorithm, providing accuracy bounds on the spectral estimates given by the quadrature algorithm~\citep{ubaru2017fast,adams2018estimating,chen2021analysis,chen2022randomized,lin2016approximating}. In neural networks, the first applications of SLQ were in vision models, where several papers reported that Hessians of trained classification tend to separate into a bulk and outlier eigenvalues, roughly equal to the number of classes ~\citep{papyan2018full,papyan2019measurements,papyan2020traces, ghorbani2019investigation}, an effect related to "neural collapse"~\citep{papyan2020prevalence}. Several open-source toolkits implement SLQ for neural network spectra, including the Tensorflow/JAX accompanying~\citet{ghorbani2019investigation}, PyHessian~\citep{yao2020pyhessian}, GradVis~\citep{chatzimichailidis2019gradvis}, and the Deep Curvature suite~\citep{granziol2019deep}. More recently, empirical spectral measurements have been used to derive optimization prescriptions in neural network training.~\citet{granziol2022learning} use SLQ to derive an on-the-fly hyperparameter learner from the Hessian spectra during training. In transformers,~\citet{zhang2024transformers} found that the Hessian spectrum is very heterogenous across parameter blocks, and further used these findings to design Adam-mini, a more memory efficient version of Adam~\citep{zhang2025adam}. To the best of our knowledge, the first application of SLQ at up to 10B parameter scale for LLMs are~\citet{granziol2025hessformer, granziol2026hessian}, running SLQ at a depth of around $10-20$ probes, which gives a poor resolution in the spectrum tails. Finally, there is a wide body of literature established on the connections between the neural tangent kernel (NTK) eigenvalues, and the Gauss-Newton ~\citep{noci2024learning,lauditi2026spectral,jiang2026understanding,kalra2025universal}.
\paragraph{Linear Stability.} The dynamics of SGD on a quadratic are a stochastic linear dynamical system. In the least squares setting with batch size $1$,~\citet{defossez2015averaged} showed that the expected loss diverges when the step size exceeds a critical threshold determined by the fourth moment of the data. ~\citet{jain2018parallelizing, ma2018power} later extended this to larger batch sizes and showed that the maximal stable learning rate grows linearly in the batch size up to a critical batch size. Motivated by the fact that at a minimizer, the loss landscape is locally convex, ~\citet{wu2018sgd,ma2021linear,wu2022alignment,velikanov2022view,mulayoff2024exact} applied this stability analysis to describe which minima can be attractors for SGD. Linear stability in neural networks has also been extensively studied in the full batch setting under the name \emph{edge of stability}. \citet{cohen2021gradient} showed that the largest Hessian eigenvalue typically increases during GD training until it reaches the critical stability threshold $2/\eta$ after which it oscillates for the remainder of training around this value. This extends empirical observations made by ~\citet{jastrzkebski2018relation,jastrzebski2020break}.~\citet{cohen2022adaptive} also studied the edge of stability phenomenon for adaptive optimizers like Adam and showed that the right notion of local stability is given by the eigenvalues of the preconditioned Hessian. Several mechanisms have been proposed for how training is sustained at the edge of stability including the catapult effect for large initial learning rates ~\citep{lewkowycz2020large} a self-stabilization effect resulting from the cubic term in the loss Taylor expansion~\citep{damian2022self}, and progressive sharpening driven by normalization~\citep{arora2022understanding}. More recently,~\citet{cohen2024understanding} have proposed an ODE capturing the dynamics of GD and adaptive optimizers at the edge of stability which empirically captures the true training dynamics. In the case of noise, a separate line of work has analyzed the dynamics of SGD and momentum at the edge of stability~\citep{andreyev2024edge,andreyev2026momentum}.

Most closely related to our stability experiments in \Cref{sec:edge_of_stability}, concurrent work by \citet{cai2026does} studies the edge of stability phenomenon in LLMs at scales up to 1.7B parameters. They probe the edge of stability by perturbing the learning rate at a fixed batch size, and observe that large batch training occurs at the edge of stability while training well below the critical batch size does not. However, because their probes only perturb the learning rate, they cannot distinguish instability caused by discretization from instability caused by stochasticity. Jointly perturbing $(\eta, B)$ allows us to separate the two, and we find that at small batch sizes training occurs at a stochastic rather than deterministic edge of stability. Our setting also differs in that we use the linearized model and the corresponding Gauss-Newton matrices as our convex quadratic proxy, while \citet{cai2026does} instead add a proximal L2 penalty to the local problem which shifts the negative eigenvalues up to zero.

\paragraph{Prescriptive Optimization Theory.} Lastly, we acknowledge two complementary lines of work striving to connect optimization theory with training practice, providing theoretically grounded prescriptions for hyperparameter selection in pretraining. The first shows that convergence bounds from convex optimization theory can be used to derive practical prescriptions for hyperparameter scaling in LLMs~\citep{schaipp2025surprising,hagele2024scaling,shulgin2026deriving,defazio2023optimal,defazio2026schedulefree+,defazio2024road}. The second models the discrete optimizer dynamics through stochastic differential equations, yielding scaling rules for hyperparameters across batch sizes and principled comparisons between adaptive methods~\citep{li2017stochastic,malladi2022sdes,compagnoni2024adaptive,compagnoni2026interaction,compagnoni2023sde,zhang2026beyond}.

\section{Quadratics are Predictive for Pretraining}
\label{sec:quadratics_are_predictive}
Taylor's theorem is a central tool for studying smooth functions, as it allows complicated nonconvex objectives to be approximated locally by polynomials, which are often far easier to analyze. Studying arbitrary nonconvex objectives directly is often too broad to yield sharp predictions for real world training: worst-case analyses have to account for pathological functions that may be irrelevant to the training runs observed in practice. Taylor's theorem provides a natural way to make the problem local. Around a point on the training trajectory, a complicated objective can be approximated by a quadratic model whose behavior is governed by the local gradient and curvature. This turns the question from one about arbitrary nonconvex optimization into a concrete empirical test: do local quadratic expansions during pretraining accurately predict the complete training dynamics? 

\subsection{Local models}
\label{sec:local_models}

Throughout the manuscript, we consider three related losses, suppressing the dependence on the minibatch for notational simplicity. Let $\ell$ be cross entropy loss, $\theta$ be the parameters, and $f_\theta$ be the model on a single sample so the loss is $\mathcal{L}(\theta) = \ell(f_\theta)$. Our first simplified loss, \texttt{prox}, linearizes $f_\theta$ around a reference point $\theta_\mathrm{ref}$ but maintains the cross-entropy objective:
\[
\mathcal{L}_{\mathrm{prox}}(\theta)
:=
\ell(f^\mathrm{lin}_\theta) \qq{where} f^\mathrm{lin}_{\theta} = f_{\theta_\mathrm{ref}} + \nabla_\theta f_{\theta_\mathrm{ref}} (\theta - \theta_\mathrm{ref}).
\]
This simplification, also known as the NTK approximation~\citep{jacot2018neural,chizat2019lazy} or the GN-proximal loss~\citep{burke1985descent,drusvyatskiy2017proximal,abreu2025potential}, reduces the non-convex dynamics to solving a convex multi-class logistic regression problem where the samples are given by the network Jacobians at $\theta_\mathrm{ref}$.

We can further simplify by Taylor expanding the proximal loss to second order to get \texttt{quad}:
\[
\mathcal{L}_{\mathrm{quad}}(\theta)
:= \mathcal{L}(\theta_\mathrm{ref}) + \nabla \mathcal{L}(\theta_{\mathrm{ref}}) (\theta - \theta_\mathrm{ref}) + \tfrac{1}{2} (\theta - \theta_\mathrm{ref})^T G (\theta - \theta_\mathrm{ref})
\]
where $G = \nabla^2 \mathcal{L}_{\mathrm{prox}}(\theta_\mathrm{ref})$ is the Gauss-Newton matrix at $\theta_\mathrm{ref}$.

Note that $\mathcal{L}_\mathrm{quad}(\cdot)$ is not the usual second order Taylor expansion, as we use the Gauss-Newton matrix $G = \nabla^2 \mathcal{L}_\mathrm{prox}$, rather than the Hessian $H = \nabla^2 \mathcal{L}$. This corresponds to keeping the positive semidefinite (PSD) term in the common Hessian decomposition:
\begin{align*}
    H(\theta) \quad=\quad \underbrace{G(\theta)}_{\succeq 0} \quad+\quad \cancel{\nabla_f \ell(f_\theta)^T \nabla^2_\theta f_\theta}.
\end{align*}
Keeping only the Gauss--Newton term ensures that the quadratic objective is convex: its curvature matrix is positive semidefinite, so the local proxy cannot become unbounded below along negative-curvature directions. Thus, our quadratic model should be viewed as the second-order approximation induced by the linearized network, rather than the full Taylor expansion of the original model. We compare the Hessian and Gauss--Newton spectra in Section~\ref{sec:spectrum}, and leave a more detailed study of the dropped term to future work.
\subsection{Definitions and Experimental Methodology}
\label{sec:methodology}

We train a 150M (non-embedding) parameter model --- 168M with embedding and unembedding matrices --- on 3B tokens from FineWeb at batch sizes $B=1, 64, 1024$, sequence length $1024$, at the optimally tuned learning rates for each batch size (see \Cref{fig:eta_B_pretraining_sweep} for the full sweep). We train our models using Adam with no weight decay, warmup of $10\%$, $\epsilon = 10^{-8}$, $\beta_1 = 0.9$ and we dynamically set $\beta_2$ at step $t$ to $\max(0.95,1 - \frac{1}{0.01t})$, so the second moment buffer $\nu$ averages over the last 1\% of tokens. For evaluation, we also track exponential moving averages (EMA) of both the parameters and the second moment buffer with parameter $\gamma_t = \max(0,1-\frac{1}{0.04t})$ which averages over the last 4\% of tokens. For cosine, we decay to $10\%$ of the peak learning rate.

For the architecture, we use a standard OLMO-style transformer with embedding dimension $1024$, depth $12$, $16$ heads. We also use the CompleteP~\citep{dey2026don} parameterization to set the per-coordinate learning rates for Adam. To ensure our experiments are not dominated by embedding layers, we set our vocab size to $8192$ by training our own byte-pair encoding (BPE) tokenizers. This ensures that the embedding and unembedding layers take up 10\% of the total parameters, which matches the ratio for many larger models~\citep{liu2024deepseek}.\footnote{Using the standard GPT-2 vocab size of 50257 would cause the embedding and unembedding layers to take up 40\% of the total parameters, which is extremely unrealistic at larger scale.} We note that with CompleteP multipliers, the learning rates appear much larger than those standard in the literature. This is because the intermediate layers in the model have a multiplier of $1/D$ over the base learning rate, where $D = 1024$. Thus $\eta = 1$ with CompleteP approximately corresponds to $\eta = 0.001$ under standard parameterization.

\subsection{Pretraining as a Chain of Local Quadratic Problems}
\begin{table}[!t]
    \centering
    \small
    \setlength{\tabcolsep}{3.7pt}
    \renewcommand{\arraystretch}{1.12}

    \begin{tabular}{@{}ccrrrrrrrrrr@{}}
        \toprule
        & & \multicolumn{10}{c}{Starting checkpoint} \\
        \cmidrule(lr){3-12}
        $B$ & Approximation
        & $10\%$ & $20\%$ & $30\%$ & $40\%$ & $50\%$
        & $60\%$ & $70\%$ & $80\%$ & $90\%$ & $100\%$ \\
        \midrule

        \multirow{2}{*}{$1$}
        & \Prox & 0.0 & 0.0 & 0.3 & 0.6 & 1.0 & 1.6 & 4.5 & 9.0 & 10.0 & 10.0 \\
        & \Quad & 0.0 & 0.0 & 0.0 & 0.3 & 0.0 & 0.3 & 1.0 & 5.5 & 9.0 & 10.0 \\
        \addlinespace[2pt]

        \rowcolor{batchshade}
        & \Prox & 0.0 & 0.0 & 0.0 & 0.3 & 1.0 & 1.9 & 3.5 & 7.1 & 10.0 & 9.4 \\
        \rowcolor{batchshade}
        \multirow{-2}{*}{$64$}
        & \Quad & 0.0 & 0.3 & 0.0 & 0.0 & 0.0 & 0.3 & 1.9 & 5.2 & 8.4 & 9.4 \\
        \addlinespace[2pt]

        \multirow{2}{*}{$1024$}
        & \Prox & 0.3 & 0.3 & 1.6 & 1.9 & 2.3 & 3.9 & 7.7 & 7.4 & 10.0 & 10.0 \\
        & \Quad & 0.3 & 0.3 & 0.3 & 1.3 & 1.9 & 2.6 & 4.8 & 6.5 & 10.0 & 5.8 \\
        \bottomrule
    \end{tabular}
    \vspace{0.5em}
    \caption{
    Percentage of the full training token budget for which the linearized loss predicts the true loss decrease from the starting checkpoint to within \(10\%\), i.e.,
    \(\left|(\hat{\mathcal{L}}_t-\mathcal{L}_t)/
    (\mathcal{L}_{\mathrm{ref}}-\mathcal{L}_t)\right|\leq 0.1\).
    Each continuation spans \(10\%\) of training, so the maximum entry is \(10\).
    }
    \label{tab:linearization-accuracy}
\end{table}

For each pretraining run, we save checkpoints $\theta_k$ at every $10\%$ of the total token budget. Starting from each $\theta_k$, we continue training for the next $10\%$ of the original budget, but replace the loss with its local expansion around $\theta_\mathrm{ref} = \theta_k$: the optimizer is run on $\lprox$ for the \texttt{prox} experiments and on $\lquad$ for the \texttt{quad} experiments. To isolate the effect of the loss approximation, everything else is kept identical to the original run: Adam is initialized from the optimizer state at checkpoint $k$, we follow the original learning rate schedule, and we maintain the same $4\%$ parameter EMA used for validation. After this continuation, we evaluate the resulting EMA parameters on the \emph{true} nonlinear model and report the full validation loss $\mathcal{L}(\cdot)$.

We show our results in \Cref{fig:cosine_linearization_loss_plots} for cosine decay, and in \Cref{fig:constant_linearization_loss_plots} (\Cref{app:constant_learning_rate}) for a constant learning rate with EMA. While early time linearizations quickly diverge from the loss, middle to late time expansions match the ground truth for roughly $5$--$10\%$ of the total training budget. In general, \texttt{prox} matches better than \texttt{quad} at early times as well, across all three batch sizes, and \texttt{quad} only begins to match after $60\%$ of training. We leave a further study of why this is the case to future work.

\section{The Hessian of LLMs}
\label{sec:hessian_llms}

In \Cref{sec:quadratics_are_predictive}, we showed that quadratic Taylor expansions are surprisingly predictive over short timescales. This is incredibly powerful, because we have a broad toolbox for studying and understanding optimization on quadratics. Importantly, these are not \emph{arbitrary} quadratic objectives, and instead possess special structure inherited from the local loss landscape. In this section, we will shed light on this structure by estimating properties of the population Hessian, which is arguably the most important quantity for quadratic optimization.

Because Adam is a preconditioned method, we distinguish between the \emph{raw} Hessian $H$ and the \emph{preconditioned} Hessian $H_P := P^{-1} H$ where $P := \mathrm{diag}[\sqrt{\nu}+\epsilon]$ is the Adam preconditioner.\footnote{We also bake in the CompleteP~\citep{dey2026don} multipliers into both the raw Hessian and preconditioned Hessian so that the contribution of different layers to the Hessian spectrum remains stable across scales.} Note that the preconditioned Hessian has real eigenvalues as $H_P$ is similar to $P^{-1/2} H P^{-1/2}$, which is symmetric. The raw Hessian governs the behavior of SGD, while the preconditioned Hessian governs the behavior of preconditioned methods like Adam \citep{cohen2022adaptive}.

We also distinguish between the Hessian of the linearized loss, the Gauss-Newton matrix $G$, and the Hessian of the original loss, the loss Hessian $H$. We will use $G,H$ to refer to the raw Gauss-Newton and Hessian matrices, and use $G_P, H_P$ to refer to the preconditioned versions.

\subsection{Methodology}

\begin{algorithm}[t]
\caption{Lanczos quadrature with reorthogonalization}
\label{alg:lanczos_quadrature}
\begin{algorithmic}[1]
\Require Symmetric matrix-vector product $v \mapsto Hv$, probe $v_0$, depth $m$
\State $q_1 \gets v_0 / \norm{v_0}$
\For{$j = 1,\ldots,m$}
    \State $z \gets H q_j$
    \State $\alpha_j \gets q_j^\top z$
    \For{$i = 1,\ldots,j$}
        \State $z \gets z - q_i(q_i^\top z)$
    \EndFor
    \State $\beta_j \gets \norm{z}$
    \State $q_{j+1} \gets z / \beta_j$
\EndFor
\State $T_m \gets \operatorname{tridiag}(\beta_1,\ldots,\beta_{m-1}; \alpha_1,\ldots,\alpha_m)$
\State Diagonalize $T_m = U \mathrm{diag}(\vartheta) U^\top$
\State $\omega \gets U_{1,:}^{\odot 2}$, $V \gets Q U$ where $Q=[q_1,\ldots,q_m]$
\State \Return Ritz values, weights and vectors $\{(\vartheta_i,\omega_i, V_{:,i})\}_{i=1}^m$
\end{algorithmic}
\end{algorithm}

\begin{algorithm}[t]
\caption{Gauss--Radau quadrature}
\label{alg:gauss_radau_quadrature}
\begin{algorithmic}[1]
\Require Tridiagonal Lanczos matrix $T_m$, final residual $\beta_m$, dimension $n$, target value $\lambda \not \in \mathrm{spec}(T_m)$
\State $\gamma \gets \lambda+\beta_m^2 e_m^\top (T_m-\lambda I)^{-1}e_m$
\State $T \gets \begin{pmatrix} T_m & \beta_m e_m \\ \beta_m e_m^\top & \gamma \end{pmatrix}$
\State $T = U\operatorname{diag}(\vartheta)U^\top$
\State $w_k \gets nU_{1k}^2$ for $k=1,\ldots,m+1$
\State $i_-(\lambda) \gets \sum_{k:\vartheta_k>\lambda} w_k$
\State $i_+(\lambda) \gets i_-(\lambda)+w_j$ where $\vartheta_j = \lambda$
\State \Return interval $[i_-(\lambda),i_+(\lambda)]$
\end{algorithmic}
\end{algorithm}

Naively, the Hessian spectrum is impossible to compute because a model with $p$ parameters has a Hessian $H \in \R^{p \times p}$, which is far too large to store or eigen-decompose. However, we can efficiently compute Hessian-vector products $v \to Hv$ which allows us to sketch the Hessian and estimate its spectrum using Lanczos quadrature. For details, we refer the reader to~\citet{golub2009matrices}, but we describe the basic idea here. Given a starting vector $v_0$ and access to matrix-vector products $v \to Hv$, Lanczos sketches the distribution $\mu(v_0)$ where $\mu(v)$ is defined by:
\begin{align*}
    \mu(v) := \sum_{i=1}^p (v \cdot u_i)^2 \delta_{\lambda_i}
\end{align*}
and $\{(\lambda_i,u_i)\}$ are the eigenvalues and eigenvectors of $H$. This can be interpreted as a reweighted empirical spectral distribution in which each eigenvalue $\lambda_i$ is given probability $(v \cdot u_i)^2$.

Using $m$ Hessian vector products, the Lanczos recurrence returns $m$ Ritz values $\vartheta_1,\ldots,\vartheta_m$ and $m$ weights $\omega_1,\ldots,\omega_m$ which sum to $1$, and produces the estimate:
\begin{align*}
    \mu(v_0) \approx \sum_{i=1}^m \omega_i \delta_{\vartheta_i}.
\end{align*}
This approximation is guaranteed to match up to the first $2m-1$ moments. An imprecise but useful interpretation is that each Ritz value $\vartheta_i$ is a representative for a group of eigenvalues $\{\lambda_j\}_{j \in \mathcal{I}}$, and the corresponding weight $\omega_i \approx \sum_{j \in \mathcal{I}} (v_0 \cdot u_i)^2$ is the total weight of this group of eigenvalues.

In this section we use the Lanczos recurrence to inspect three properties of the Hessian:
\begin{enumerate}
    \item \textbf{Spectral Decay:} estimate the map $i \to \lambda_i$ to look for power laws in the spectrum
    \item \textbf{Eigenvector Distribution:} estimate the map $i \to \norm{\Pi_\ell u_i}^2$ where $\Pi_\ell$ denotes the orthogonal projection onto layers of type $\ell$ (\texttt{embed}, \texttt{unembed}, \texttt{attn.q/k/v}, \texttt{mlp.up/down})
    \item \textbf{Gradient-Hessian Alignment:} estimate the map $i \to (\nabla \mathcal{L}(\theta) \cdot u_i)^2$
\end{enumerate}

\paragraph{Experimental Details} For (1) and (2) we use Lanczos probes with random $v_0 \sim S^{d-1}$ so that $\E_{v_0} \mu(v_0)$ is the empirical spectral distribution of $H$. For (3) we run an additional gradient-start probe with $v_0 = \nabla \mathcal{L}(\theta)$. For each of the checkpoints $B=1,64,1024$ and $T=10\%,50\%,100\%$ we run these Lanczos probes on the preconditioned Gauss Newton $G_P$, the preconditioned Hessian $H_P$, the raw Gauss Newton $G$, and the raw Hessian $H$. All Lanczos probes were run with a probe depth of $m=1200$ and stored the full Lanczos basis with two-pass modified Gram–Schmidt reorthogonalization in fp32 precision. The matrix-vector products were computed over 10M tokens.

\paragraph{Accuracy and Ablations} These large scale Lanczos runs require a lot of memory and compute, so we ran ablations to test which parameters were most important:
\begin{itemize}
    \item \textbf{Probe count:} We empirically found that a single random probe was sufficient to resolve the spectrum $i \to \lambda_i$ to high accuracy, and given a fixed matrix-vector product budget, the primary bottleneck is always the Lanczos depth $m$, so we ran one random probe and one gradient probe for each (checkpoint, matrix) combination. This is justified by the fact that deep in the tails, the weight of each Ritz value is standing in for the average weight of a large block of eigenvalues, which creates an averaging effect even with a single probe.
    \item \textbf{Precision:} We found that the Lanczos three term recurrence quickly lost accuracy around depth $100$ regardless of precision, so we elected to store the full Lanczos basis and do multi-pass reorthogonalization for all runs. We found that storing the basis in fp16 led to a drop in accuracy, but there was no drop in accuracy between fp32 and fp64, so we ran in fp32. The basis for a single checkpoint was around 750GB in memory which we sharded across 16 H100 GPUs.
    \item \textbf{Sample Size:} We compared the spectrum at 10M tokens and 100M tokens and found no significant difference, even deep in the tails, so we elected to compute the matrix-vector products using 10M tokens for compute efficiency (Figure~\ref{fig:100m_comparison}).
\end{itemize}

We also tested our methodology against synthetic spectra for which the ground truth is known. These comparisons can be found in \Cref{fig:spectrum_bootstrap}. We found extremely strong agreement between the SLQ estimates and the ground truth spectra, even deep into the tails.

\subsection{The Hessian Spectrum}\label{sec:spectrum}

\begin{figure}
    \includegraphics[width=\columnwidth]{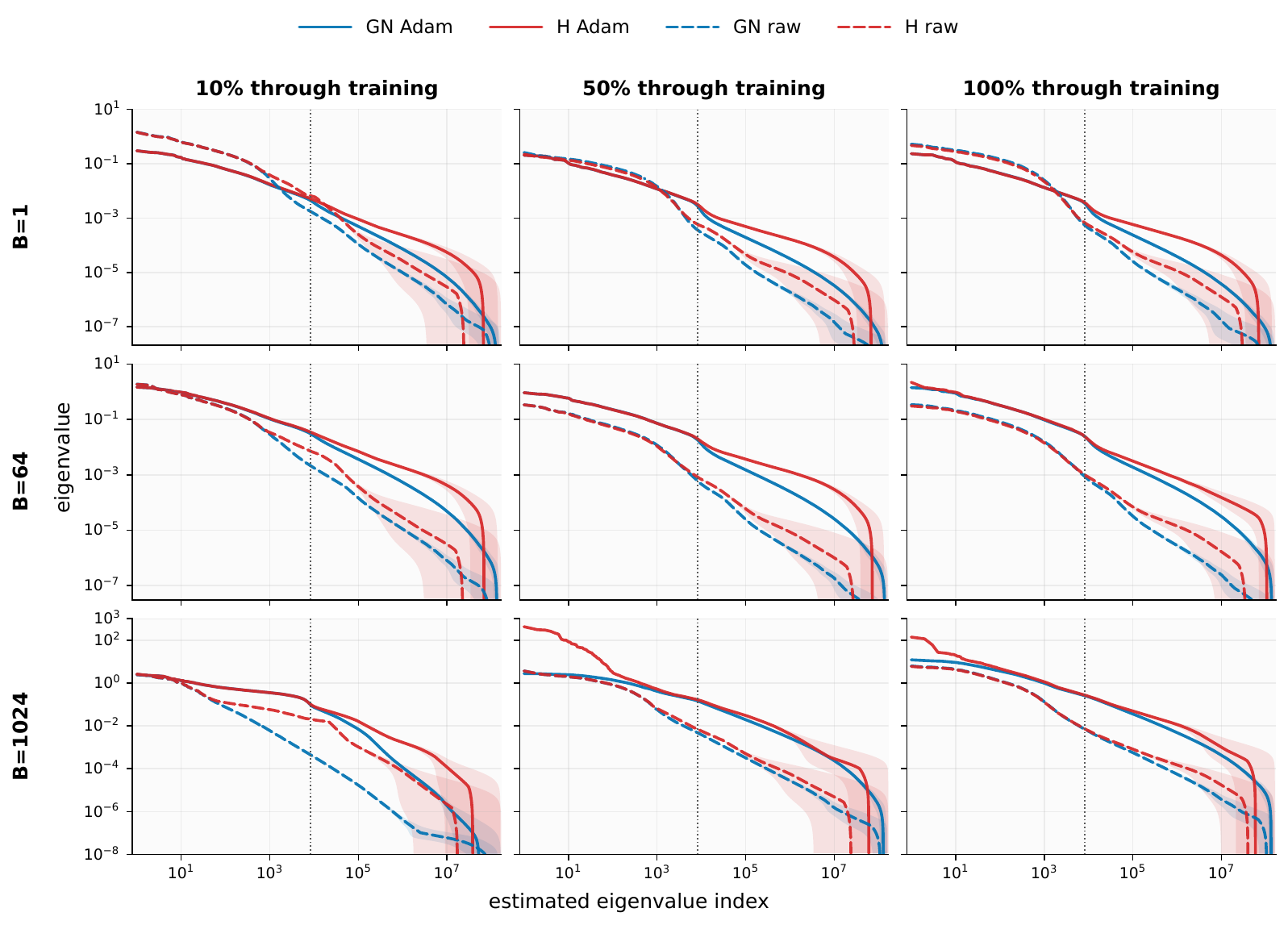}
    \caption{We estimate the spectrum of the Gauss Newton matrix, the Hessian, and their preconditioned versions at different checkpoints. The solid lines denote the best estimate for the spectrum, and the shaded bands represent error bands from finite depth Lanczos probes ($m=1200$).}
    \label{fig:cosine_spectra}
\end{figure}

\begin{figure}[t]
  \centering

  \begin{minipage}[t][0.373\textheight][t]{0.29\linewidth}
    \vspace{0pt}
    \centering
    \includegraphics[width=\linewidth]{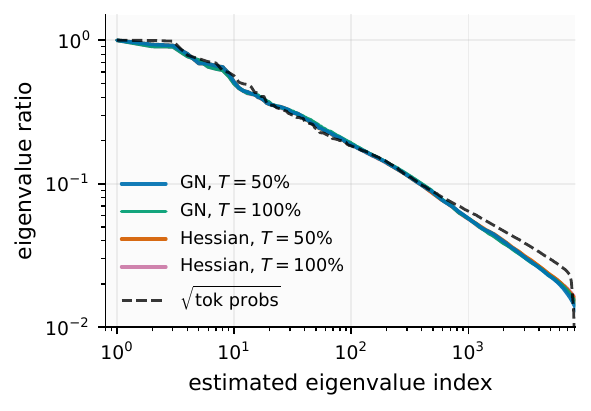}
    \subcaption{Top eigenvalues at $B=1$}
    \label{fig:tok-head}

    \vfill

    \includegraphics[width=\linewidth]{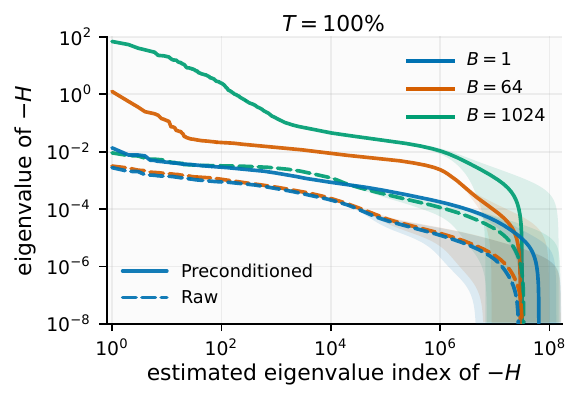}
    \subcaption{Negative Hessian eigenvalues}
    \label{fig:negative-evals}
  \end{minipage}
  \hfill
  \begin{minipage}[t]{0.34\linewidth}
    \vspace{0pt}
    \centering
    \includegraphics[width=\linewidth]{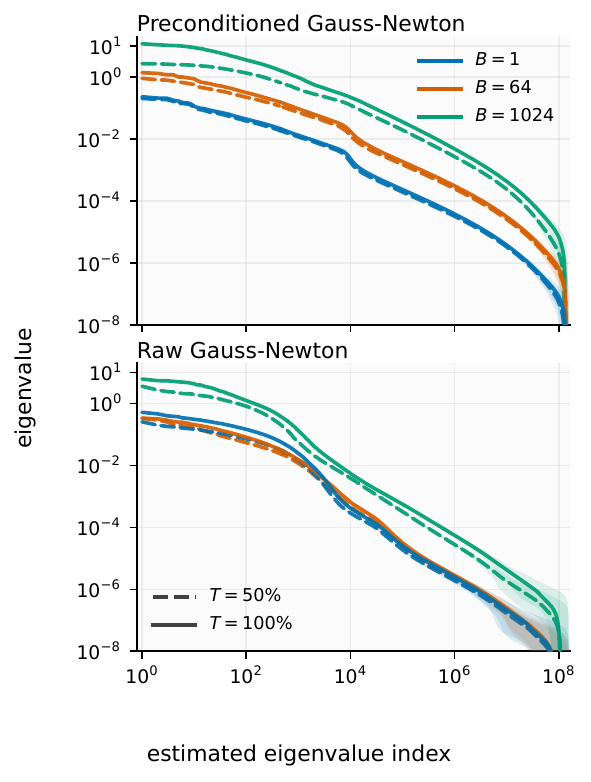}
    \subcaption{Spectra at $T=50\%,100\%$}
    \label{fig:spectrum-training}
  \end{minipage}
  \hfill
  \begin{minipage}[t]{0.31\linewidth}
    \vspace{0pt}
    \centering
    \includegraphics[width=\linewidth]{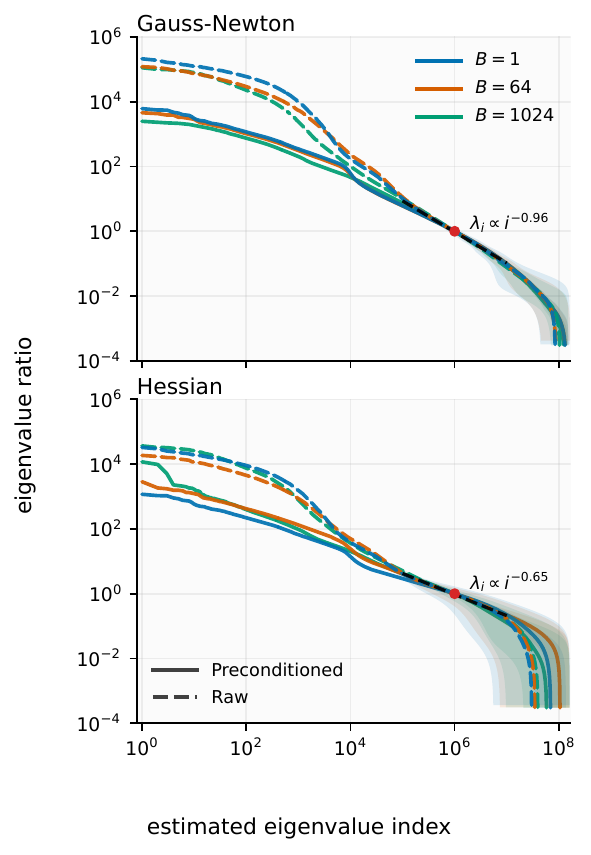}
    \subcaption{Final spectra aligned at $i=10^6$}
    \label{fig:pinned-spectrum}
  \end{minipage}

  \caption{Analyses of the Hessian spectra for cosine decay. See \Cref{fig:constant_spectrum} for the constant schedule.}
  \label{fig:spectrum-summary}
\end{figure}

We plot the spectra for the runs using cosine decay in \Cref{fig:cosine_spectra}. We begin with two important points about plotting methodology and expected accuracy before interpreting the experimental results.

\paragraph{Plotting Methodology:} Lanczos naturally estimates the map $i \to \lambda_i$. Specifically, the Lanczos recurrence guarantees that if $i^\star(\lambda)$ is the index of $\lambda$ in the empirical spectral density $\mu(v_0)$ then $i^\star(\vartheta_j) \in [i_{j-1},i_j]$ where $i_j := p \sum_{k=1}^j \omega_k$ are the partial sums of the Ritz weights. This guarantee can be extended to intermediate $\lambda \ne \vartheta_j$ using Gauss-Radau bounds (\Cref{alg:gauss_radau_quadrature}). These guarantee that for any $\lambda$, $i^\star(\lambda) \in [i_-(\lambda),i_+(\lambda)]$. We can therefore create a uniform grid in log-space for $\lambda$, compute $i_{\pm}(\lambda)$, and shade the resulting band. We also plot a solid best fit curve for visualization purposes.

\paragraph{Theoretical Error Estimates:} A simple heuristic is that Lanczos can resolve the spectrum with small multiplicative error for eigenvalues $\lambda \gtrsim \lambda_1/m^2$. For $m=1200$, this means we can resolve eigenvalues up to 1.5M times smaller than the top eigenvalue, which is consistent with the Gauss-Radau error bands in our experiments. We empirically found that the error bands for the Hessian are larger than for the Gauss Newton, which we attribute to the fact that Lanczos has to resolve both the positive and negative extremal eigenvalues before resolving the tail near $\lambda = 0$, which effectively halves its effective probe depth $m$.

\subsubsection{Experimental Results}

We observed remarkable structure in these spectra, and highlight a few key points:

\paragraph{Head-Tail Split:} In most of the Hessian/Gauss Newton spectra, especially at smaller batch sizes, there is a sharp transition around $i = 8192$, which is our model's vocab size. In \Cref{fig:evecs_cosine}, we show that this is not a coincidence as the top $V$ eigenvectors are dominated by the $D \times V$ unembedding layer. We believe this may be related to neural collapse~\citep{papyan2018full,papyan2019measurements,papyan2020prevalence,papyan2020traces}, in which each token $v$ has a unique unembedding vector $h_v \in S^{D-1}(\sqrt{D})$, and the Hessian of the last layer is well approximated by the rank $V$ matrix: $\sum_{v \in [V]} p_v (h_v e_v^T)^{\otimes 2}$ where $p_v$ are the marginal token probabilities. To test this, we plotted $\sqrt{p_v}$ against the top $V$ eigenvalues of the
preconditioned spectrum\footnote{The square root is to account for Adam's
preconditioner at small batch sizes} and found a surprisingly good fit. For
example, both the true spectrum and the token probabilities $p_v$ have 3 nearly equal
outliers (corresponding to the tokens \tok{the}, \tok{,}, and \tok{.}).

\paragraph{Spectrum Evolution:} We observe that both the preconditioned and raw Hessian and Gauss Newton matrices do not change significantly between the $50\%$ and $100\%$ checkpoints. We fit the capacity exponent as well in~\Cref{fig:spectrum-training} and see that in the tail we get roughly $\alpha \approx 0.96$ for Gauss-Newton and $\alpha \approx 0.65$ for Hessian. Note that as shown in Figure~\ref{fig:pinned-spectrum}, this exponent, when fitted on the tail, does not depend on batch size or whether we take into account the preconditioner.

\paragraph{Gauss-Newton Hessian agreement:} We observe that at small and medium batch sizes, the spectra of the $G_P$ and $H_P$ agree best in the head, while at large batches they agree best in the tails. Similarly, the spectra of the raw matrices $G,H$ always agree in the head and agree best in the tail for $B=1024$.

\paragraph{Tail Universality:} Perhaps our most surprising finding is that regardless of the batch size or preconditioner, the tail of the spectrum appears to be \emph{\textbf{universal}}, meaning that the power law is identical and can only be shifted by an absolute constant. To visualize this, we aligned the spectra of both the raw and preconditioned Gauss Newton and Hessian matrices at $i=10^6$. Especially for the Gauss Newton, the tail for $i \in [10^5,10^7]$ appears to match for all batch sizes and regardless of whether the matrix is preconditioned, implying the Adam preconditioner does not affect the tail.

\paragraph{Negative Eigenvalues:} Unsurprisingly, both the preconditioned Hessian and raw Hessian have negative eigenvalues, shown in~\Cref{fig:negative-evals}. Perhaps more surprising is that the negative eigenvalues of the preconditioned Hessian $H_P$ are as large as the positive ones at $B=64$ and significantly larger at $B=1024$. In \Cref{fig:full_evec_cosine,fig:full_evec_constant} we can see that the corresponding eigenvectors are almost entirely localized to \texttt{mlp.up}, the MLP layer that reads from the residual stream. We hypothesize that these are related to dead neurons which actuate the rare token effect (see \Cref{par:rare_events}).

\subsection{The Hessian Eigenvectors}

\begin{figure}[H]
    \includegraphics[width=\columnwidth]{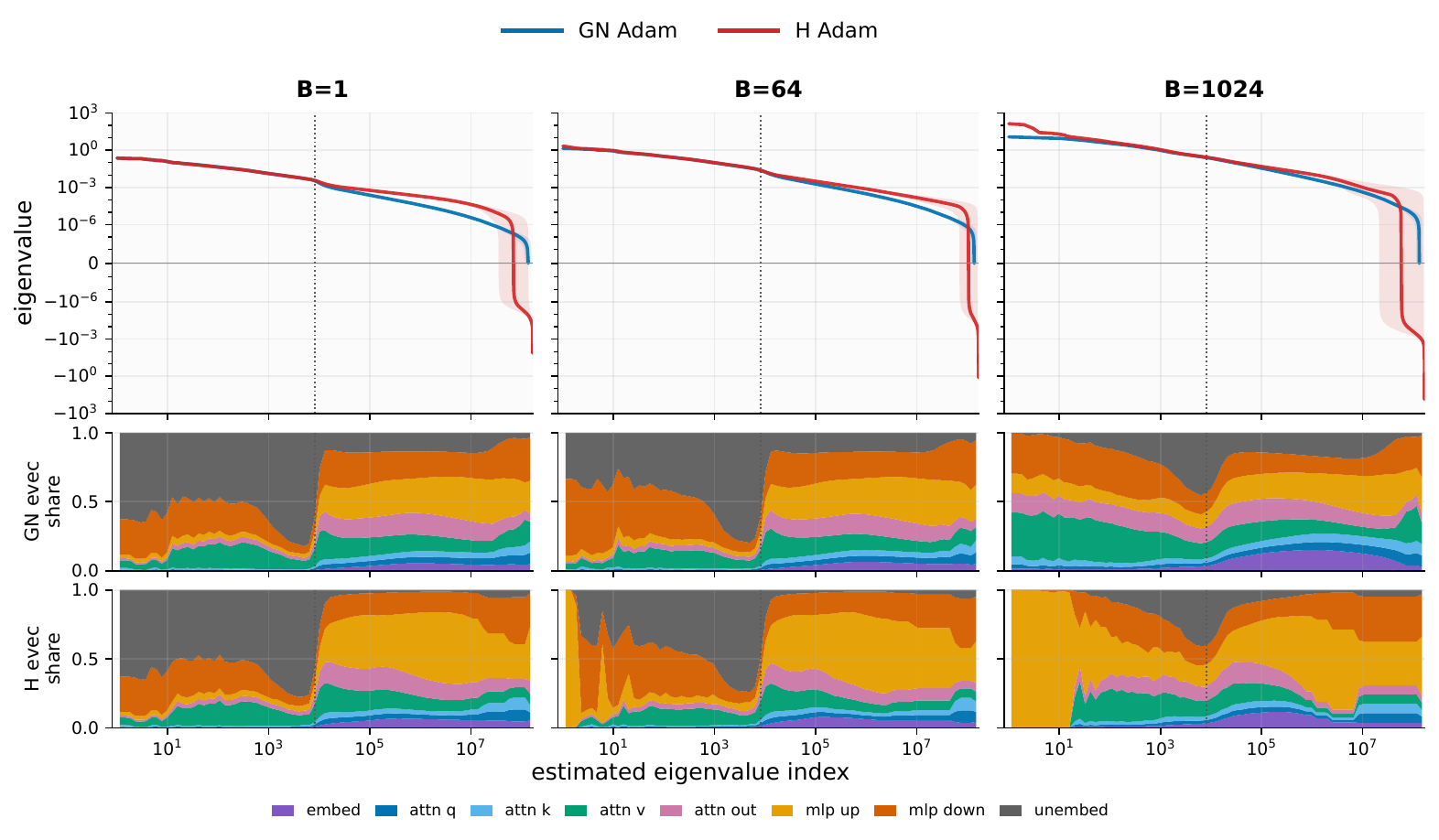}
    \caption{An estimate of the mass of each eigenvector in each parameter group using SLQ.}
    \label{fig:evecs_cosine}
\end{figure}

Because we maintain the basis $V$ in \Cref{alg:lanczos_quadrature}, we can also compute the Ritz vectors $u_1,\ldots,u_m$ associated to the Ritz values $\vartheta_1,\ldots,\vartheta_m$. This lets us estimate which parts of the network are contributing to which parts of the Hessian spectrum. Explicitly, given a unit vector $v$ and a parameter block $\mathcal{I}$, we define the mass of $v$ in $\mathcal{I}$ to be $\sum_{i \in \mathcal{I}} v_i^2$. Note that if we partition the parameters into disjoint subsets $\{I_1,\ldots,I_k\}$ then these masses must sum to $1$. We plot the results in \Cref{fig:evecs_cosine}.

We highlight three main observations:

\paragraph{The head is dominated by \texttt{unembed}:} In \Cref{sec:spectrum}, we observed that for $B=1,64$, there was a clear split between the head and tail of the spectrum around $i = V$. In \Cref{fig:evecs_cosine}, we can see that the corresponding eigenvectors live largely in the unembedding layer, which we connected to neural collapse \citep{papyan2020prevalence} in \Cref{sec:spectrum}. This effect disappears for $B = 1024$. We attribute this to the fact that under neural collapse the Hessian of the unembedding layer is adapted to the standard coordinate basis, so full batch Adam is able to completely precondition it away \citep{cohen2024understanding}, while small batch Adam only applies a square root.

\paragraph{The tail is spread uniformly across the layers:} After the head, the remaining eigenvectors appear to be spread evenly across the different types of layers, and the eigenvector mass is approximately proportional to the number of parameters in each layer.

\paragraph{Large Hessian outliers are dominated by \texttt{mlp.up}:} We observed in \Cref{sec:spectrum} that at $B=1024$, the Hessian has extreme outlier eigenvalues, both positive and negative. In \Cref{fig:evecs_cosine} (for a more complete plot across batch sizes and checkpoints, we refer the reader to \Cref{fig:full_evec_cosine} for cosine and \Cref{fig:full_evec_constant} for constant) we can see that the corresponding eigenvectors live entirely in \texttt{mlp.up}. Our hypothesis is that nearly dead neurons in \texttt{mlp.up} cause the Adam second moment vector $\nu$ to collapse to $0$, so that $\mathrm{diag}(\nu^{-1/2}) H$ diverges in those directions. See \Cref{par:rare_events} for additional discussion.

\subsection{Source Conditions (Hessian-gradient alignment)}

\begin{figure}[H]
    \includegraphics[width=\columnwidth]{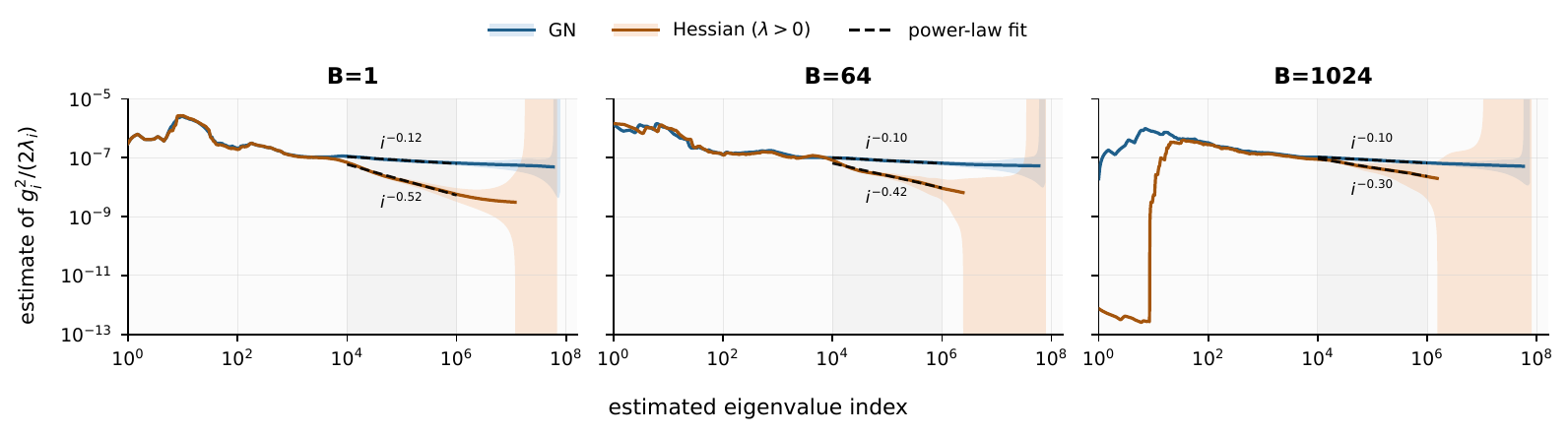}
    \caption{An estimate of the source condition $\ev{u_i,\nabla L(\theta)}^2/\lambda_i$ using SLQ at the final 100\% checkpoint.}
    \label{fig:source_cosine}
\end{figure}

In addition to a random Lanczos probe, we also run Lanczos starting from the gradient, which captures the distribution of the gradient across the spectrum. In analyses of linear regression, this is often called the \emph{source} condition and is parameterized by the exponent $\beta$:
\begin{align}
\ev{u_i, \nabla \mathcal{L}(\theta)}^2/\lambda_i \propto i^{-\beta} \tag{Source Condition} \\
\lambda_i \propto i^{-\alpha} \tag{Capacity Condition} 
\end{align}
where $\{(\lambda_i,u_i)\}$ are the eigenvalues and eigenvectors of the Hessian. This can also be interpreted as the amount of excess loss contained in the $i$-th eigenvector because for a quadratic objective:
\begin{align}
    \mathcal{L}(\theta) - \mathcal{L}^\star = \sum_{i=1}^p \frac{\ev{u_i, \nabla \mathcal{L}(\theta)}^2}{2\lambda_i}. \label{eq:source_L*}
\end{align}
This easily generalizes to preconditioned methods. If we define the symmetrized preconditioned Hessian to be $P^{-1/2} H P^{-1/2}$ and replace the gradient with the semi-preconditioned gradient $g \gets P^{-1/2} g$, then the same decomposition holds. In \Cref{fig:source_cosine}, we estimate this source condition for both the preconditioned Gauss-Newton matrix and the preconditioned Hessian. We note that the map that Lanczos naturally estimates is the cumulative sum $i \to \sum_{j < i} (g \cdot u_j)^2$, so to estimate $i \to (g \cdot u_j)^2$, we have to differentiate this map which is numerically unstable. We therefore first average the value over a multiplicative window of $[i/2,2i]$ to smooth it before differentiating.

The primary observation is that in the part of the tail we can resolve, the source condition for the Gauss Newton matrix is consistently around $\beta \in [0.1,0.2]$ while the source condition for the Hessian is closer to $\beta \in [0.3,0.6]$. This contradicts the common assumption $\beta > 1$ in the literature on power laws, which guarantees \cref{eq:source_L*} is summable as $p \to \infty$ and, in the standard bias-variance decomposition, gives rise to power laws in the bias which depend on the source and capacity exponents. Instead, when $\beta < 1$, as we measure here, the problem actually behaves like it is finite dimensional ~\citep[Figure 2]{paquette20244}. When $\beta < 1$, there is a critical time $T^\star = p^\alpha$, where $p$ is the number of parameters, after which the bias decays exponentially. Meanwhile the variance term hits a noise floor without averaging and decays like $O(1/T)$ with averaging~\citep{polyak1992acceleration,dieuleveut2020bridging}. We empirically verify this in \Cref{fig:joint_power_law_fit} by fitting a power law of the form $\mathcal{L}_t = \mathcal{L}^\star + AT^{-\gamma}$ to all of our optimally tuned runs where $\mathcal{L}^\star \approx 2.502$ is shared. Nearly all runs have $\gamma \approx 1$, which matches the variance dominated regime predicted by theory.

\section{Edge of Stability}
\label{sec:edge_of_stability}

\citet{cohen2021gradient} demonstrated that full-batch optimization typically occurs at the \emph{edge of stability} (EOS), meaning that increasing the learning rate to $(1+\epsilon) \eta$ would cause the optimizer to diverge on the local quadratic Taylor expansion, while decreasing it to $(1-\epsilon) \eta$ would cause the optimizer to converge. For gradient descent, this is captured by condition $\eta \approx 2/\lambda_1(H)$. This was later generalized to adaptive optimizers, including Adam, in~\citet{cohen2022adaptive} where they showed that if you freeze the preconditioner, adaptive optimizers also typically operate at the edge of stability.

These empirical results primarily hold in the full-batch (or at least very large batch) setting. However, a similar phenomenon occurs at smaller batch sizes. In particular, there is a \emph{stochastic edge of stability} regime in which decreasing your batch size can cause you to diverge on the local quadratic Taylor expansion. In this section, we empirically test whether optimization in LLMs occurs at either the deterministic or stochastic edge of stability, and we compare these results with the theoretical predictions made by linear stability analysis.

\subsection{Methodology}

To test whether optimization occurs at the edge of stability, we reuse a similar experimental setup as in Section~\ref{sec:methodology}. We Taylor expand the model using \texttt{quad}, around the EMA iterate $\bar{\theta}$, so that the objective on each minibatch is quadratic. We then freeze Adam's second moment buffer $\nu$ to the EMA $\bar{\nu}$. We then run preconditioned SGD with this fixed preconditioner for $10\%$ of the total token budget, or until the loss diverges. To test how close a checkpoint is to the edge of stability, we run this experiment over a $5 \times 5$ grid of learning rate and batch size multipliers: $\{1/4,1/2,1,2,4\}$.\footnote{For batch size $1$ we could not use a batch size multiplier less than $1$ so we generated a $5 \times 3$ grid.} We also repeat this experiment over $10$ random seeds, where each random seed shuffles the sequences in the dataset. We say a run diverges if its loss hits $100$, and we plot the fraction of runs which diverge as a function of the learning rate and batch size multipliers in Figure~\ref{fig:eos_cosine_grid}. While we arbitrarily cap the loss divergence thresholds at $100$, \Cref{fig:eos_cosine_grid} is incredibly stable for any cutoff between $10$ and $10^5$.

\paragraph{Extreme Outliers} An important detail, which results from freezing the preconditioner, is that the \texttt{quad} training is now susceptible to divergence due to extreme outliers in the dynamics, which we observed were connected to rare tokens or dead neurons~\citep{voita2024neurons}. We believe an illustrative example is to look at the dynamics of the embedding matrix $E$. In every minibatch, if token $i$ does not occur then its corresponding embedding $E_i$ gets no gradient. Because the second moment buffer $\nu$ is an EMA of these zero gradients, if the token isn't seen for roughly $\tfrac{1}{1 - \beta_2}$ steps, then those entries of $\nu$ become exponentially small. If this token later reoccurs, then Adam will divide its gradient by $\sqrt{\nu}+\epsilon \approx 0$, which can cause a spike. While this phenomenon is clearest for the embedding matrix, it can also occur throughout the model, e.g. from neurons which are rarely activated~\citep{voita2024neurons}. To circumvent this issue, we estimate the Frobenius norm of the preconditioned Gauss-Newton matrix for each sequence $\|P^{-1} G_i\|_F^2$ and filter out the top $1\%$ outliers. Without this correction, nearly all hyperparameters diverged after a single outsized spike.\label{par:rare_events}

Before we discuss the experimental results, we will briefly discuss the predictions that theory makes for stability of SGD on quadratics so that we can better interpret the experimental results and validate the predictive power of the theory.

\subsection{Theoretical Predictions for Stability on Quadratics}\label{sec:edge_of_stability:theory}

\begin{figure}[!htp]
    \centering
    \includegraphics[width=0.6\linewidth]{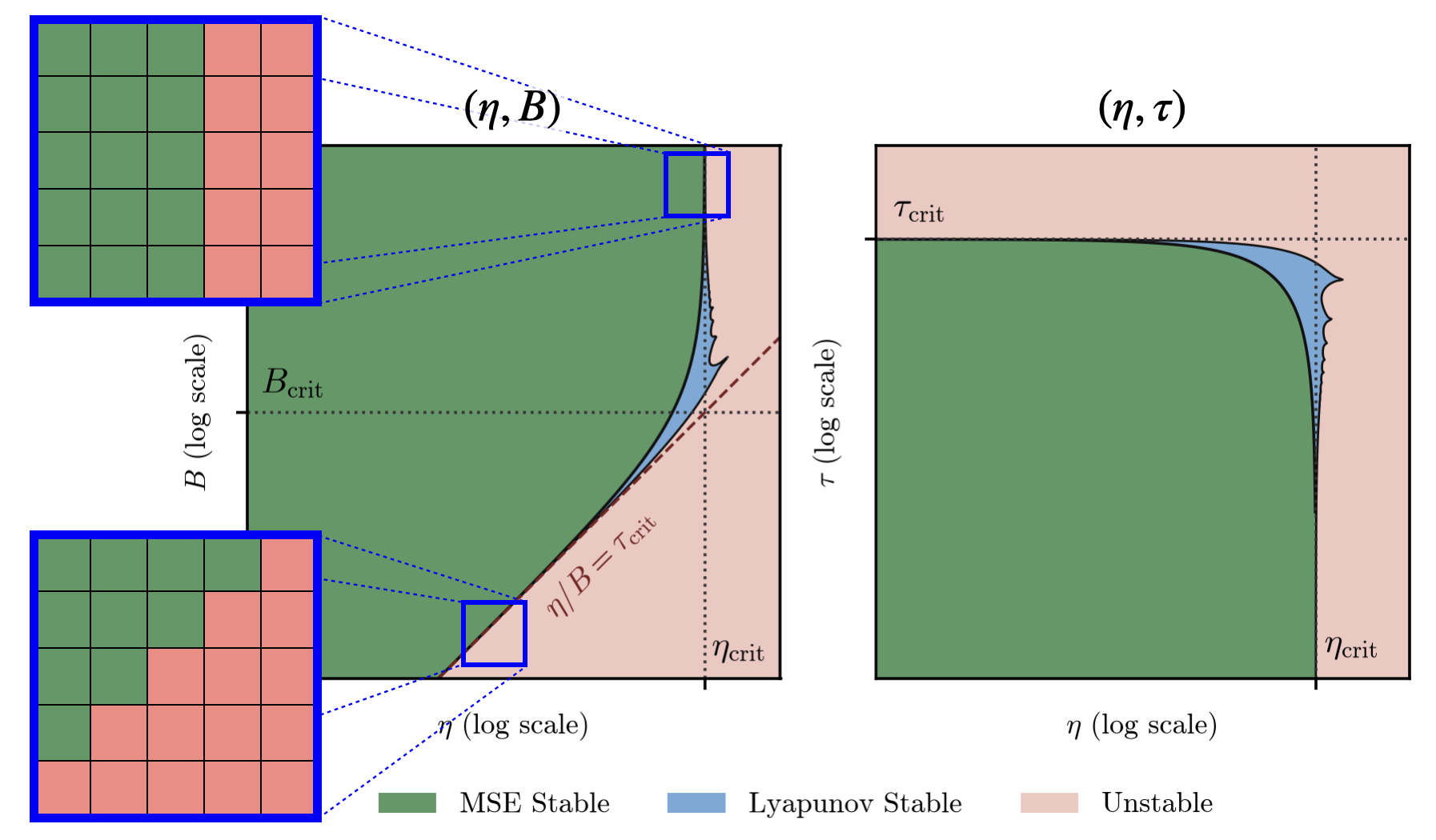}
    \caption{A toy simulation of the linear stability on a 1D regression problem where $H_i \stackrel{i.i.d.}{\sim} \mathrm{Poisson}(B_\text{crit})$. Note the MSE stable region is monotonic in $\eta,B,\tau$ while the Lyapunov stable region is not. Our EOS experiments determine where on this cartoon training occurs by perturbing $\eta,B$.}
    \label{fig:stable_region}
\end{figure}

\begin{figure}[!htp]
    \centering
    \includegraphics[width=1.0\linewidth]{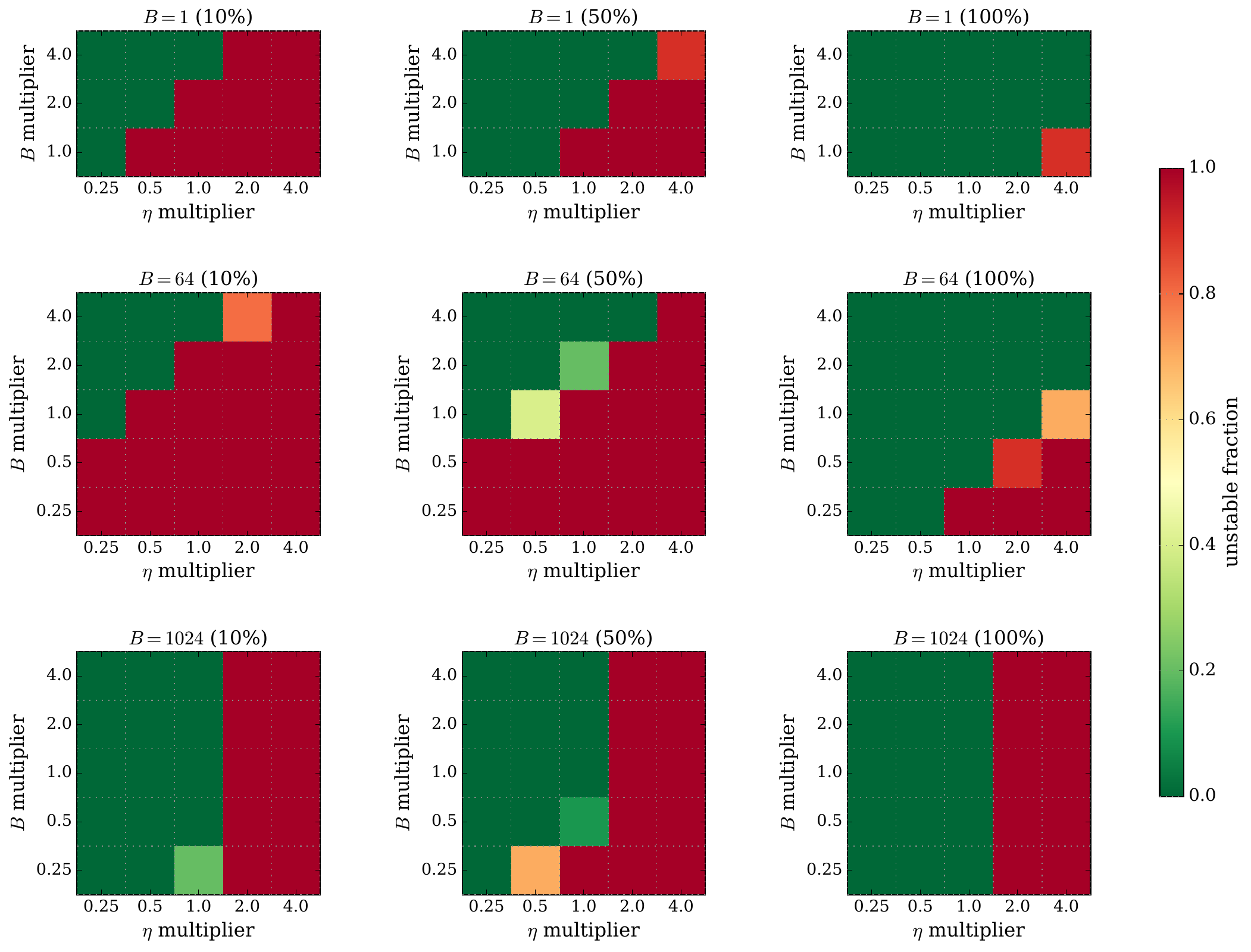}
    \caption{Heatmap showing instability for models trained with cosine decay as a function of $(\eta, B)$ across batch size $B=1$, $64$ and $1024$, at $10\%$, $50\%$ and $100\%$ of training. Cell color indicates how many seeds diverged out of a total of $10$ seeds per run. We provide the analogous figure for constant learning rate in Figure~\ref{fig:eos_constant_grid} (Appendix~\ref{app:constant_learning_rate}).}
    \label{fig:eos_cosine_grid}
\end{figure}

The dynamics of SGD on a quadratic are a linear dynamical system. In particular, if $H_\mathcal{B}$ denotes the Hessian on a minibatch $\mathcal{B}$ then we can rewrite the SGD update as:
\begin{equation}
\begin{aligned}
    \theta - \theta^\star
    &\gets \theta - \theta^\star - \eta \nabla \mathcal{L}_\mathcal{B}(\theta)
    && \why{SGD update} \\
    &= \theta - \theta^\star
    - \eta [\nabla \mathcal{L}_\mathcal{B}(\theta) - \nabla \mathcal{L}_\mathcal{B}(\theta^\star)]
    - \eta \nabla \mathcal{L}_\mathcal{B}(\theta^\star)
    && \why{add and subtract} \\
    &= \underbrace{(I - \eta H_\mathcal{B})}_{\mathclap{\text{stochastic contraction}}}
    (\theta - \theta^\star)
    - \underbrace{\eta \nabla \mathcal{L}_\mathcal{B}(\theta^\star)}_{\text{additive noise}}.
    && \why{gradients are affine}
\end{aligned}\label{eq:sgd_linear_update}
\end{equation}
In other words, at every step we multiply the current displacement to $\theta^\star$ by a random matrix $I - \eta H_\mathcal{B}$ then add a random noise vector independent of $\theta$. For linear dynamical systems of this form, there are two relevant notions of stability: mean square stability and Lyapunov stability.
\begin{definition}
    A linear dynamical system $\theta_t$ is \textbf{mean square} stable if $\lim_{t \to \infty } \mathbb{E}[\|\theta_t\|^2] < \infty$, and \textbf{Lyapunov} stable if $\lim_{t \to \infty} \E \log \|\theta_t\| < \infty$.
\end{definition}
Note that by Jensen's inequality, mean square stability implies Lyapunov stability:
$$
\mathbb{E}[\log \|\theta\|] = \tfrac{1}{2} \mathbb{E}[\log \|\theta\|^2] \le \tfrac{1}{2} \log \E \|\theta\|^2.
$$
For most processes, mean-square stability and Lyapunov stability nearly coincide. However, mean square stability is significantly more sensitive to heavy tails. As an illustrative example, let $\eta = B = 1$ and run SGD on the loss functions $\mathcal{L}_1 = \tfrac{1}{2} \theta^2$, $\mathcal{L}_2 = \tfrac{3}{2} \theta^2$. Then at every step you multiply $\theta$ by either $1 - \eta = 0$ or $1 - 3\eta = -2$, each with probability $1/2$. The marginal at time $t$ is $0$ with probability $1-2^{-t}$ and $(-2)^t$ with probability $2^{-t}$. This process is mean square unstable because $\mathbb{E}[\theta^2] = 2^t \to \infty$, but it is Lyapunov stable, which reflects that $\theta_t \to 0$ almost surely. Mean-square stability captures wild fluctuations, while Lyapunov stability captures the ``typical'' behavior of the process. We will begin by briefly discussing mean-square stability before discussing Lyapunov stability, as it is more mathematically tractable and provides better intuition.

\paragraph{Mean Square Stability} There is a simple closed form for mean-square stability. If
$$
\Sigma = \E[(\theta - \theta^\star)(\theta - \theta^\star)^\top]
$$
then from \cref{eq:sgd_linear_update}, $\Sigma$ is updated by
\begin{align*}
    \Sigma \gets \mathbb{E}_\mathcal{B}[(I - \eta H_\mathcal{B}) \Sigma (I - \eta H_\mathcal{B})] + \eta^2 C \qq{where} C = \Cov_\mathcal{B}[\nabla \mathcal{L}_\mathcal{B}(\theta^\star)].
\end{align*}
This is a deterministic linear system for $\Sigma$, so $\Sigma$ will either converge or diverge depending on the eigenvalues of the linear operator $\Phi_{\eta,B}$:
\begin{align*}
    \Phi_{\eta,B}[\Sigma] := \mathbb{E}_\mathcal{B}[(I - \eta H_\mathcal{B}) \Sigma (I - \eta H_\mathcal{B})].
\end{align*}
Stability of the $\Sigma$ recurrence requires $\lambda_1(\Phi_{\eta,B}) \le 1$. Because $\E \|\theta_t\|^2 = \mathrm{tr}~\Sigma_t$, this is also the condition for mean square stability. Note that the learning rate $\eta$ enters directly, and the batch size $B$ enters implicitly through the sampling of the minibatch $\mathcal{B}$. With some effort, it is possible to make these dependencies more explicit. For the remainder of this section, we will use $\tau := \eta/B$ to denote the ``temperature'' or learning rate to batch size ratio of the process.
\begin{proposition}[{\citealp[Theorem 5 \& Proposition 6]{mulayoff2024exact}}]~
Let $\tau := \eta/B$ be the temperature and let $\mathcal{S}_\text{MSE} = \{(\eta,\tau) ~|~ \lambda_1(\Phi_{\eta,B})<1\}$ be the set of mean square stable hyperparameters for SGD. Then the stable set $\mathcal{S}_\text{MSE}$ satisfies the following properties:
\begin{itemize}
    \item $\mathcal{S}_\text{MSE}$ is monotonic in $\eta$ and $\tau$, i.e. if $(\eta,\tau) \in \mathcal{S}_\text{MSE}$ then $(c_1 \eta,c_2\tau) \in \mathcal{S}_\text{MSE}$ for any $c_1,c_2 \le 1$.
    \item There is a maximum learning rate $\eta_\text{crit}$ and maximum temperature $\tau_\text{crit}$, i.e. $\mathcal{S}_\text{MSE} \subseteq [0,\eta_\text{crit}) \times [0,\tau_\text{crit})$. $\eta_\text{crit}$ is given by $2/\lambda_1(H)$, as for GD, while $\tau_\text{crit}$ depends on the homogeneity of the minibatch Hessians.
\end{itemize}
\end{proposition}
The effect of these observations is that in $\eta,\tau$ space, the set of hyperparameters for which SGD is mean-square stable is a rounded rectangle with edges $\eta < \eta_\text{crit}$ and $\tau < \tau_\text{crit}$. Transforming this into $\eta,B$ space, this becomes a wedge shape. This is reflected in the toy simulation in \Cref{fig:stable_region}.

\paragraph{Lyapunov Stability} Lyapunov stability is significantly less structured than mean square stability. For example, \Cref{fig:stable_region} shows a linear regression problem where $(\eta,B)$ is Lyapunov unstable but $(\eta,2B)$ is stable, so Lyapunov stability does not satisfy a monotonicity property. Furthermore, the set of Lyapunov stable hyperparameters is often not even connected. This distinction turns out to be relevant for LLMs. In our experiments, we did observe settings in which $(\eta,B)$ is stable but $(\eta/2,B/2)$ is not stable, which violates monotonicity. This is inconsistent with mean-square stability and suggests that there are regimes during training in which SGD is Lyapunov stable but not mean square stable.

\subsection{Experimental Results}
Our experimental results demonstrate that the majority of training occurs within a factor of $2$ of the edge of stability, and the shape of the plots is incredibly consistent with the theoretical predictions in \Cref{sec:edge_of_stability:theory}. Figure~\ref{fig:eos_cosine_grid} shows that for $B=1$ and $B=64$, the stability boundary follows the diagonal line, whereas at $B=1024$ there is a flat cutoff, consistent with the illustration from \Cref{fig:stable_region}. Moreover, while the $(\eta, B) = (1, 1)$ cell is not always in the stability region, the $(0.5, 1)$ cell is always stable, indicating that every model is running at most a factor of $2$ over the stability boundary, which is typical even for full-batch gradient descent \citep{cohen2021gradient} and matches the conclusions of \citet{cai2026does}.

Our results also contextualize the finding of \citet{cai2026does} that the edge of stability disappears at small batch sizes. One confounder is that their probes run for a fixed number of steps ($\le 50$), rather than a fixed number of tokens, so at small batch sizes the simulations process very few tokens and can fail to detect instabilities. Our results suggest that after normalizing by tokens, rather than steps, LLMs operate within a factor of $2$ of the stochastic edge of stability at small batch sizes.

\section{Discussion and Limitations}

\paragraph{Quadratics as a Pretraining Sandbox.} While LLM loss landscapes are not \textit{globally} quadratic, the optimizer appears to enter regions that are well approximated by a prox-linear or Gauss-Newton Taylor expansion for a nontrivial fraction of the training budget (up to 10\%). This may partially explain why intuition and algorithms for the quadratic model seem to extend to LLM pretraining.

\paragraph{Structure of the Local Quadratic.} When the local dynamics are captured by the \texttt{quad} Taylor expansion, we can understand these dynamics through the spectrum of the Hessian/Gauss-Newton matrix. We estimated these spectra using deep Lanczos quadrature probes and found that the local quadratic models have extremely structured spectra, with power-law like behavior in both the head and the tail. We also find that the tails are \emph{universal} meaning that their shape does not vary with training time or batch size and is not affected by the Adam preconditioner. The SLQ probes also allow us to localize different parts of the spectrum to different layers in the network.

\paragraph{Local Stability.} Motivated by the observation that full-batch training typically occurs at the edge of stability~\citep{cohen2021gradient,cohen2022adaptive} and by theoretical analyses of linear stability of SGD ~\citep{mulayoff2024exact,defossez2015averaged}, we empirically test whether LLM pretraining operates near a (stochastic) stability boundary. We first study the set of locally stable hyperparameters $(\eta,B)$, and show how the shape of its boundary determines whether the instability is caused by discretization or stochasticity. We then locate where training occurs within this set by perturbing $(\eta, B)$ and checking for instability. At small batch sizes, training typically occurs at a stochastic edge of stability, while at large batch sizes it typically occurs at a deterministic edge of stability.

\paragraph{Progressive Sharpening and Curvature Regularization.} It has been widely observed that the stochasticity in SGD/Adam drives optimization towards flatter regions of the loss landscape \cite{keskar2017on, jastrzkebski2018relation}. This curvature regularization effect cannot be captured by the quadratic model since the curvature is by definition fixed. Therefore, the fact that the quadratic model does seem to capture the training dynamics near the end of the training suggests that this curvature regularization effect is relatively weak by the end of training. However, we do observe evidence of progressive sharpening for our runs with cosine decay. As the learning rate decays, the runs leave the edge of stability and the curvature once again begins to grow, which matches the observations in \citet{cai2026does}.

\paragraph{Limitations.} Our experiments focus on a fixed model scale, architecture, and optimizer. While the agreement between the local and ground truth model is remarkable, it is an important future direction to explore how the model size and other optimizers --- particularly second order optimizers --- influence the spectrum exponents and loss agreement. Moreover, Figure~\ref{fig:cosine_linearization_loss_plots} shows that early on in training, both the quadratic and the prox-linear approximations are much less accurate than at mid-to-late time, suggesting that higher-order terms in the loss play an important role in the early phase of pretraining. Finally, while the late time agreement for cosine is remarkable, constant with EMA tracks the original very poorly (Figure~\ref{fig:constant_linearization_loss_plots}). We speculate that this is related to the edge of stability behavior of the constant runs during training and we leave a further study to future work.

\subsection{Conclusion}
Overall, our findings support a simple but powerful mental model: LLM pretraining can be locally understood as optimization over a sequence of structured quadratic problems. Instead of asking for a single global theory of a highly nonconvex loss landscape, we can ask how the local quadratic model changes along the trajectory, how the optimizer interacts with its spectrum, and how we can leverage these insights to design more efficient training pipelines. In this sense, the quadratic model is not merely a toy problem, but an experimentally validated proxy for the optimization dynamics of pretraining, and a useful foundation for building theory that applies to real world models.

\label{sec:discussion}

\subsection*{Acknowledgements and Funding}
The authors
would like to thank Max Shad and Bala Desinghu for their help with the cluster. AM, DM,
PN acknowledge the support of a Kempner Institute Graduate Research Fellowship. CP is supported by an NSF CAREER Award (IIS-2239780), DARPA grants
DIAL-FP-038 and AIQ-HR00112520041, the Simons Collaboration on the Physics of Learning and
Neural Computation, and the William F. Milton Fund from Harvard University. The authors acknowledge that this work has been made possible in part by a gift from the Chan
Zuckerberg Initiative Foundation to establish the Kempner Institute for the Study of Natural and
Artificial Intelligence. SK and DM acknowledge support from the Office of Naval Research under
award N0001422-1-2377 and the National Science Foundation Grant under award \#IIS 2229881. DM
is also supported by a Simons Investigator Fellowship, NSF grant DMS-2134157, DARPA grant
W911NF2010021, and DOE grant DE-SC0022199.

\bibliographystyle{plainnatyear}
\bibliography{references}

@article{burke1985descent,
  title={Descent methods for composite nondifferentiable optimization problems},
  author={Burke, James V},
  journal={Mathematical Programming},
  volume={33},
  number={3},
  pages={260--279},
  year={1985},
  publisher={Springer}
}

@article{drusvyatskiy2017proximal,
  title={The proximal point method revisited},
  author={Drusvyatskiy, Dmitriy},
  journal={arXiv preprint arXiv:1712.06038},
  year={2017}
}

@article{abreu2025potential,
  title={The potential of second-order optimization for llms: A study with full gauss-newton},
  author={Abreu, Natalie and Vyas, Nikhil and Kakade, Sham and Morwani, Depen},
  journal={arXiv preprint arXiv:2510.09378},
  year={2025}
}

@article{cohen2021gradient,
  title={Gradient descent on neural networks typically occurs at the edge of stability},
  author={Cohen, Jeremy M and Kaur, Simran and Li, Yuanzhi and Kolter, J Zico and Talwalkar, Ameet},
  journal={arXiv preprint arXiv:2103.00065},
  year={2021}
}

@article{cohen2022adaptive,
  title={Adaptive gradient methods at the edge of stability},
  author={Cohen, Jeremy M and Ghorbani, Behrooz and Krishnan, Shankar and Agarwal, Naman and Medapati, Sourabh and Badura, Michal and Suo, Daniel and Cardoze, David and Nado, Zachary and Dahl, George E and others},
  journal={arXiv preprint arXiv:2207.14484},
  year={2022}
}

@article{polyak1964some,
  title={Some methods of speeding up the convergence of iteration methods},
  author={Polyak, Boris T},
  journal={Ussr computational mathematics and mathematical physics},
  volume={4},
  number={5},
  pages={1--17},
  year={1964},
  publisher={Elsevier}
}

@article{zhang2024does,
  title={How Does Critical Batch Size Scale in Pre-training?},
  author={Zhang, Hanlin and Morwani, Depen and Vyas, Nikhil and Wu, Jingfeng and Zou, Difan and Ghai, Udaya and Foster, Dean and Kakade, Sham},
  journal={arXiv preprint arXiv:2410.21676},
  year={2024}
}

@article{polyak1992acceleration,
  title={Acceleration of stochastic approximation by averaging},
  author={Polyak, Boris T and Juditsky, Anatoli B},
  journal={SIAM journal on control and optimization},
  volume={30},
  number={4},
  pages={838--855},
  year={1992},
  publisher={SIAM}
}

@article{meterez2025seesaw,
  title={Seesaw: Accelerating Training by Balancing Learning Rate and Batch Size Scheduling},
  author={Meterez, Alexandru and Morwani, Depen and Wu, Jingfeng and Oncescu, Costin-Andrei and Pehlevan, Cengiz and Kakade, Sham},
  journal={arXiv preprint arXiv:2510.14717},
  year={2025}
}

@inproceedings{kidambi2018insufficiency,
  title={On the insufficiency of existing momentum schemes for stochastic optimization},
  author={Kidambi, Rahul and Netrapalli, Praneeth and Jain, Prateek and Kakade, Sham},
  booktitle={2018 Information Theory and Applications Workshop (ITA)},
  pages={1--9},
  year={2018},
  organization={IEEE}
}

@article{cohen2024understanding,
  title={Understanding optimization in deep learning with central flows},
  author={Cohen, Jeremy M and Damian, Alex and Talwalkar, Ameet and Kolter, J Zico and Lee, Jason D},
  journal={arXiv preprint arXiv:2410.24206},
  year={2024}
}

@article{meterez2026anytime,
  title={Anytime Pretraining: Horizon-Free Learning-Rate Schedules with Weight Averaging},
  author={Meterez, Alexandru and Nair, Pranav Ajit and Morwani, Depen and Pehlevan, Cengiz and Kakade, Sham},
  journal={arXiv preprint arXiv:2602.03702},
  year={2026}
}

@article{morwani2026compute,
  title={Compute Efficiency and Serial Runtime Tradeoffs for Stochastic Momentum Methods},
  author={Morwani, Depen and Meterez, Alexandru and Nair, Pranav and Kakade, Sham},
  journal={arXiv preprint arXiv:2606.19179},
  year={2026}
}

@inproceedings{zou2021benign,
  title={Benign overfitting of constant-stepsize sgd for linear regression},
  author={Zou, Difan and Wu, Jingfeng and Braverman, Vladimir and Gu, Quanquan and Kakade, Sham},
  booktitle={Conference on Learning Theory},
  pages={4633--4635},
  year={2021},
  organization={PMLR}
}

@article{lanczos1950iteration,
  title={An iteration method for the solution of the eigenvalue problem of linear differential and integral operators},
  author={Lanczos, Cornelius},
  journal={Journal of research of the National Bureau of Standards},
  volume={45},
  number={4},
  pages={255--282},
  year={1950}
}

@article{dey2026don,
  title={Don't be lazy: CompleteP enables compute-efficient deep transformers},
  author={Dey, Nolan and Zhang, Bin and Noci, Lorenzo and Li, Mufan and Bordelon, Blake and Bergsma, Shane and Pehlevan, Cengiz and Hanin, Boris and Hestness, Joel},
  journal={Advances in Neural Information Processing Systems},
  volume={38},
  pages={137707--137739},
  year={2026}
}

@inproceedings{voita2024neurons,
  title={Neurons in large language models: Dead, n-gram, positional},
  author={Voita, Elena and Ferrando, Javier and Nalmpantis, Christoforos},
  booktitle={Findings of the Association for Computational Linguistics: ACL 2024},
  pages={1288--1301},
  year={2024}
}

@inproceedings{chen2021analysis,
  title={Analysis of stochastic Lanczos quadrature for spectrum approximation},
  author={Chen, Tyler and Trogdon, Thomas and Ubaru, Shashanka},
  booktitle={International Conference on Machine Learning},
  pages={1728--1739},
  year={2021},
  organization={PMLR}
}

@book{golub2009matrices,
  title={Matrices, moments and quadrature with applications},
  author={Golub, Gene H and Meurant, G{\'e}rard},
  year={2009},
  publisher={Princeton University Press}
}

@article{lin2016approximating,
  title={Approximating spectral densities of large matrices},
  author={Lin, Lin and Saad, Yousef and Yang, Chao},
  journal={SIAM review},
  volume={58},
  number={1},
  pages={34--65},
  year={2016},
  publisher={SIAM}
}

@article{ubaru2017fast,
  title={Fast estimation of tr(f(A)) via stochastic Lanczos quadrature},
  author={Ubaru, Shashanka and Chen, Jie and Saad, Yousef},
  journal={SIAM Journal on Matrix Analysis and Applications},
  volume={38},
  number={4},
  pages={1075--1099},
  year={2017},
  publisher={SIAM}
}

@article{adams2018estimating,
  title={Estimating the spectral density of large implicit matrices},
  author={Adams, Ryan P and Pennington, Jeffrey and Johnson, Matthew J and Smith, Jamie and Ovadia, Yaniv and Patton, Brian and Saunderson, James},
  journal={arXiv preprint arXiv:1802.03451},
  year={2018}
}

@article{chen2022randomized,
  title={Randomized matrix-free quadrature for spectrum and spectral sum approximation},
  author={Chen, Tyler and Trogdon, Thomas and Ubaru, Shashanka},
  journal={arXiv preprint arXiv:2204.01941},
  pages={55},
  year={2022}
}

@article{sagun2016eigenvalues,
  title={Eigenvalues of the hessian in deep learning: Singularity and beyond},
  author={Sagun, Levent and Bottou, Leon and LeCun, Yann},
  journal={arXiv preprint arXiv:1611.07476},
  year={2016}
}

@article{sagun2017empirical,
  title={Empirical analysis of the hessian of over-parametrized neural networks},
  author={Sagun, Levent and Evci, Utku and Guney, V Ugur and Dauphin, Yann and Bottou, Leon},
  journal={arXiv preprint arXiv:1706.04454},
  year={2017}
}

@article{papyan2018full,
  title={The full spectrum of deepnet hessians at scale: Dynamics with sgd training and sample size},
  author={Papyan, Vardan},
  journal={arXiv preprint arXiv:1811.07062},
  year={2018}
}

@article{papyan2019measurements,
  title={Measurements of three-level hierarchical structure in the outliers in the spectrum of deepnet hessians},
  author={Papyan, Vardan},
  journal={arXiv preprint arXiv:1901.08244},
  year={2019}
}

@article{papyan2020traces,
  title={Traces of class/cross-class structure pervade deep learning spectra},
  author={Papyan, Vardan},
  journal={Journal of Machine Learning Research},
  volume={21},
  number={252},
  pages={1--64},
  year={2020}
}

@article{papyan2020prevalence,
  title={Prevalence of neural collapse during the terminal phase of deep learning training},
  author={Papyan, Vardan and Han, XY and Donoho, David L},
  journal={Proceedings of the National Academy of Sciences},
  volume={117},
  number={40},
  pages={24652--24663},
  year={2020},
  publisher={National Academy of Sciences}
}

@inproceedings{ghorbani2019investigation,
  title={An investigation into neural net optimization via hessian eigenvalue density},
  author={Ghorbani, Behrooz and Krishnan, Shankar and Xiao, Ying},
  booktitle={International Conference on Machine Learning},
  pages={2232--2241},
  year={2019},
  organization={PMLR}
}

@inproceedings{yao2020pyhessian,
  title={Pyhessian: Neural networks through the lens of the hessian},
  author={Yao, Zhewei and Gholami, Amir and Keutzer, Kurt and Mahoney, Michael W},
  booktitle={2020 IEEE international conference on big data (Big data)},
  pages={581--590},
  year={2020},
  organization={IEEE}
}

@inproceedings{chatzimichailidis2019gradvis,
  title={Gradvis: Visualization and second order analysis of optimization surfaces during the training of deep neural networks},
  author={Chatzimichailidis, Avraam and Keuper, Janis and Pfreundt, Franz-Josef and Gauger, Nicolas R},
  booktitle={2019 IEEE/ACM Workshop on Machine Learning in High Performance Computing Environments (MLHPC)},
  pages={66--74},
  year={2019},
  organization={IEEE}
}

@article{granziol2019deep,
  title={Deep curvature suite},
  author={Granziol, Diego and Wan, Xingchen and Garipov, Timur},
  journal={arXiv preprint arXiv:1912.09656},
  year={2019}
}

@article{granziol2022learning,
  title={Learning rates as a function of batch size: A random matrix theory approach to neural network training},
  author={Granziol, Diego and Zohren, Stefan and Roberts, Stephen},
  journal={Journal of Machine Learning Research},
  volume={23},
  number={173},
  pages={1--65},
  year={2022}
}

@article{zhang2024transformers,
  title={Why transformers need adam: A hessian perspective},
  author={Zhang, Yushun and Chen, Congliang and Ding, Tian and Li, Ziniu and Sun, Ruoyu and Luo, Zhi-Quan},
  journal={Advances in neural information processing systems},
  volume={37},
  pages={131786--131823},
  year={2024}
}

@inproceedings{zhang2025adam,
  title={Adam-mini: Use fewer learning rates to gain more},
  author={Zhang, Yushun and Chen, Congliang and Li, Ziniu and Ding, Tian and Wu, Chenwei and Kingma, Diederik Durk and Ye, Yinyu and Luo, Zhi-Quan and Sun, Ruoyu},
  booktitle={International Conference on Learning Representations},
  volume={2025},
  pages={28033--28063},
  year={2025}
}

@article{granziol2025hessformer,
  title={HessFormer: Hessians at Foundation Scale},
  author={Granziol, Diego},
  journal={arXiv preprint arXiv:2505.11564},
  year={2025}
}

@article{granziol2026hessian,
  title={Hessian Spectral Analysis at Foundation Model Scale},
  author={Granziol, Diego and Juarev, Khurshid},
  journal={arXiv preprint arXiv:2602.00816},
  year={2026}
}

@article{noci2024learning,
  title={Why do learning rates transfer? reconciling optimization and scaling limits for deep learning},
  author={Noci, Lorenzo and Meterez, Alexandru and Hofmann, Thomas and Orvieto, Antonio},
  journal={arXiv preprint arXiv:2402.17457},
  year={2024}
}

@article{lauditi2026spectral,
  title={Spectral Dynamics in Deep Networks: Feature Learning, Outlier Escape, and Learning Rate Transfer},
  author={Lauditi, Clarissa and Pehlevan, Cengiz and Bordelon, Blake},
  journal={arXiv preprint arXiv:2605.07870},
  year={2026}
}

@article{jiang2026understanding,
  title={Understanding the evolution of the neural tangent kernel at the edge of stability},
  author={Jiang, Kaiqi and Cohen, Jeremy and Li, Yuanzhi},
  journal={Advances in Neural Information Processing Systems},
  volume={38},
  pages={31996--32036},
  year={2026}
}

@inproceedings{kalra2025universal,
  title={Universal sharpness dynamics in neural network training: Fixed point analysis, edge of stability, and route to chaos},
  author={Kalra, Dayal Singh and He, Tianyu and Barkeshli, Maissam},
  booktitle={International Conference on Learning Representations},
  volume={2025},
  pages={55966--56000},
  year={2025}
}

@inproceedings{defossez2015averaged,
  title={Averaged least-mean-squares: Bias-variance trade-offs and optimal sampling distributions},
  author={D{\'e}fossez, Alexandre and Bach, Francis},
  booktitle={Artificial Intelligence and Statistics},
  pages={205--213},
  year={2015},
  organization={PMLR}
}

@article{jain2018parallelizing,
  title={Parallelizing stochastic gradient descent for least squares regression: mini-batching, averaging, and model misspecification},
  author={Jain, Prateek and Kakade, Sham M and Kidambi, Rahul and Netrapalli, Praneeth and Sidford, Aaron},
  journal={Journal of machine learning research},
  volume={18},
  number={223},
  pages={1--42},
  year={2018}
}

@inproceedings{ma2018power,
  title={The power of interpolation: Understanding the effectiveness of SGD in modern over-parametrized learning},
  author={Ma, Siyuan and Bassily, Raef and Belkin, Mikhail},
  booktitle={International Conference on Machine Learning},
  pages={3325--3334},
  year={2018},
  organization={PMLR}
}

@inproceedings{mulayoff2024exact,
  title={Exact mean square linear stability analysis for SGD},
  author={Mulayoff, Rotem and Michaeli, Tomer},
  booktitle={The Thirty Seventh Annual Conference on Learning Theory},
  pages={3915--3969},
  year={2024},
  organization={PMLR}
}

@article{wu2018sgd,
  title={How sgd selects the global minima in over-parameterized learning: A dynamical stability perspective},
  author={Wu, Lei and Ma, Chao and others},
  journal={Advances in Neural Information Processing Systems},
  volume={31},
  year={2018}
}

@article{ma2021linear,
  title={On linear stability of sgd and input-smoothness of neural networks},
  author={Ma, Chao and Ying, Lexing},
  journal={Advances in Neural Information Processing Systems},
  volume={34},
  pages={16805--16817},
  year={2021}
}

@article{velikanov2022view,
  title={A view of mini-batch SGD via generating functions: conditions of convergence, phase transitions, benefit from negative momenta},
  author={Velikanov, Maksim and Kuznedelev, Denis and Yarotsky, Dmitry},
  journal={arXiv preprint arXiv:2206.11124},
  year={2022}
}

@article{wu2022alignment,
  title={The alignment property of SGD noise and how it helps select flat minima: A stability analysis},
  author={Wu, Lei and Wang, Mingze and Su, Weijie},
  journal={Advances in Neural Information Processing Systems},
  volume={35},
  pages={4680--4693},
  year={2022}
}

@article{jastrzkebski2018relation,
  title={On the relation between the sharpest directions of DNN loss and the SGD step length},
  author={Jastrz{\k{e}}bski, Stanis{\l}aw and Kenton, Zachary and Ballas, Nicolas and Fischer, Asja and Bengio, Yoshua and Storkey, Amos},
  journal={arXiv preprint arXiv:1807.05031},
  year={2018}
}

@article{jastrzebski2020break,
  title={The break-even point on optimization trajectories of deep neural networks},
  author={Jastrzebski, Stanislaw and Szymczak, Maciej and Fort, Stanislav and Arpit, Devansh and Tabor, Jacek and Cho, Kyunghyun and Geras, Krzysztof},
  journal={arXiv preprint arXiv:2002.09572},
  year={2020}
}

@inproceedings{arora2022understanding,
  title={Understanding gradient descent on the edge of stability in deep learning},
  author={Arora, Sanjeev and Li, Zhiyuan and Panigrahi, Abhishek},
  booktitle={International Conference on Machine Learning},
  pages={948--1024},
  year={2022},
  organization={PMLR}
}

@article{lewkowycz2020large,
  title={The large learning rate phase of deep learning: the catapult mechanism},
  author={Lewkowycz, Aitor and Bahri, Yasaman and Dyer, Ethan and Sohl-Dickstein, Jascha and Gur-Ari, Guy},
  journal={arXiv preprint arXiv:2003.02218},
  year={2020}
}

@inproceedings{ahn2022understanding,
  title={Understanding the unstable convergence of gradient descent},
  author={Ahn, Kwangjun and Zhang, Jingzhao and Sra, Suvrit},
  booktitle={International conference on machine learning},
  pages={247--257},
  year={2022},
  organization={PMLR}
}

@article{damian2022self,
  title={Self-stabilization: The implicit bias of gradient descent at the edge of stability},
  author={Damian, Alex and Nichani, Eshaan and Lee, Jason D},
  journal={arXiv preprint arXiv:2209.15594},
  year={2022}
}

@article{jacot2018neural,
  title={Neural tangent kernel: Convergence and generalization in neural networks},
  author={Jacot, Arthur and Gabriel, Franck and Hongler, Cl{\'e}ment},
  journal={Advances in neural information processing systems},
  volume={31},
  year={2018}
}

@article{lee2019wide,
  title={Wide neural networks of any depth evolve as linear models under gradient descent},
  author={Lee, Jaehoon and Xiao, Lechao and Schoenholz, Samuel and Bahri, Yasaman and Novak, Roman and Sohl-Dickstein, Jascha and Pennington, Jeffrey},
  journal={Advances in neural information processing systems},
  volume={32},
  year={2019}
}

@article{chizat2019lazy,
  title={On lazy training in differentiable programming},
  author={Chizat, Lenaic and Oyallon, Edouard and Bach, Francis},
  journal={Advances in neural information processing systems},
  volume={32},
  year={2019}
}

@article{fort2020deep,
  title={Deep learning versus kernel learning: an empirical study of loss landscape geometry and the time evolution of the neural tangent kernel},
  author={Fort, Stanislav and Dziugaite, Gintare Karolina and Paul, Mansheej and Kharaghani, Sepideh and Roy, Daniel M and Ganguli, Surya},
  journal={Advances in Neural Information Processing Systems},
  volume={33},
  pages={5850--5861},
  year={2020}
}

@article{atanasov2021neural,
  title={Neural networks as kernel learners: The silent alignment effect},
  author={Atanasov, Alexander and Bordelon, Blake and Pehlevan, Cengiz},
  journal={arXiv preprint arXiv:2111.00034},
  year={2021}
}

@article{long2021properties,
  title={Properties of the after kernel},
  author={Long, Philip M},
  journal={arXiv preprint arXiv:2105.10585},
  year={2021}
}

@article{vyas2022limitations,
  title={Limitations of the ntk for understanding generalization in deep learning},
  author={Vyas, Nikhil and Bansal, Yamini and Nakkiran, Preetum},
  journal={arXiv preprint arXiv:2206.10012},
  year={2022}
}

@article{ortiz2021can,
  title={What can linearized neural networks actually say about generalization?},
  author={Ortiz-Jim{\'e}nez, Guillermo and Moosavi-Dezfooli, Seyed-Mohsen and Frossard, Pascal},
  journal={Advances in Neural Information Processing Systems},
  volume={34},
  pages={8998--9010},
  year={2021}
}

@article{park2023trak,
  title={Trak: Attributing model behavior at scale},
  author={Park, Sung Min and Georgiev, Kristian and Ilyas, Andrew and Leclerc, Guillaume and Madry, Aleksander},
  journal={arXiv preprint arXiv:2303.14186},
  year={2023}
}

@inproceedings{malladi2023kernel,
  title={A kernel-based view of language model fine-tuning},
  author={Malladi, Sadhika and Wettig, Alexander and Yu, Dingli and Chen, Danqi and Arora, Sanjeev},
  booktitle={International Conference on Machine Learning},
  pages={23610--23641},
  year={2023},
  organization={PMLR}
}

@article{bai2020taylorized,
  title={Taylorized training: Towards better approximation of neural network training at finite width},
  author={Bai, Yu and Krause, Ben and Wang, Huan and Xiong, Caiming and Socher, Richard},
  journal={arXiv preprint arXiv:2002.04010},
  year={2020}
}

@article{bai2019beyond,
  title={Beyond linearization: On quadratic and higher-order approximation of wide neural networks},
  author={Bai, Yu and Lee, Jason D},
  journal={arXiv preprint arXiv:1910.01619},
  year={2019}
}

@article{liu2024deepseek,
  title={Deepseek-v3 technical report},
  author={Liu, Aixin and Feng, Bei and Xue, Bing and Wang, Bingxuan and Wu, Bochao and Lu, Chengda and Zhao, Chenggang and Deng, Chengqi and Zhang, Chenyu and Ruan, Chong and others},
  journal={arXiv preprint arXiv:2412.19437},
  year={2024}
}

@article{andreyev2026momentum,
  title={Momentum further constrains sharpness at the edge of stochastic stability},
  author={Andreyev, Arseniy and Ananthkumar, Advikar and Walden, Marc and Poggio, Tomaso and Beneventano, Pierfrancesco},
  journal={arXiv preprint arXiv:2604.14108},
  year={2026}
}

@article{andreyev2024edge,
  title={Edge of stochastic stability: Revisiting the edge of stability for sgd},
  author={Andreyev, Arseniy and Beneventano, Pierfrancesco},
  journal={arXiv preprint arXiv:2412.20553},
  year={2024}
}

@article{zhang2019algorithmic,
  title={Which algorithmic choices matter at which batch sizes? insights from a noisy quadratic model},
  author={Zhang, Guodong and Li, Lala and Nado, Zachary and Martens, James and Sachdeva, Sushant and Dahl, George and Shallue, Chris and Grosse, Roger B},
  journal={Advances in neural information processing systems},
  volume={32},
  year={2019}
}

@article{paquette2021dynamics,
  title={Dynamics of stochastic momentum methods on large-scale, quadratic models},
  author={Paquette, Courtney and Paquette, Elliot},
  journal={Advances in Neural Information Processing Systems},
  volume={34},
  pages={9229--9240},
  year={2021}
}

@article{zhang2024optimality,
  title={The optimality of (accelerated) SGD for high-dimensional quadratic optimization},
  author={Zhang, Haihan and Liu, Yuanshi and Chen, Qianwen and Fang, Cong},
  journal={arXiv preprint arXiv:2409.09745},
  year={2024}
}

@article{wu2025risk,
  title={Risk comparisons in linear regression: Implicit regularization dominates explicit regularization},
  author={Wu, Jingfeng and Bartlett, Peter L and Kakade, Sham M and Lee, Jason D and Yu, Bin},
  journal={arXiv preprint arXiv:2509.17251},
  year={2025}
}

@article{ferbach2026dimension,
  title={Dimension-adapted momentum outscales SGD},
  author={Ferbach, Damien and Everett, Katie and Gidel, Gauthier and Paquette, Elliot and Paquette, Courtney},
  journal={Advances in Neural Information Processing Systems},
  volume={38},
  pages={112780--112977},
  year={2026}
}

@article{zhang2026scaling,
  title={Scaling laws for precision in high-dimensional linear regression},
  author={Zhang, Dechen and Tang, Xuan and Liang, Yingyu and Zou, Difan},
  journal={arXiv preprint arXiv:2602.19241},
  year={2026}
}

@article{varre2021last,
  title={Last iterate convergence of SGD for Least-Squares in the Interpolation regime.},
  author={Varre, Aditya Vardhan and Pillaud-Vivien, Loucas and Flammarion, Nicolas},
  journal={Advances in Neural Information Processing Systems},
  volume={34},
  pages={21581--21591},
  year={2021}
}

@inproceedings{varre2022accelerated,
  title={Accelerated sgd for non-strongly-convex least squares},
  author={Varre, Aditya and Flammarion, Nicolas},
  booktitle={Conference on Learning Theory},
  pages={2062--2126},
  year={2022},
  organization={PMLR}
}

@article{bordelon2021learning,
  title={Learning curves for SGD on structured features},
  author={Bordelon, Blake and Pehlevan, Cengiz},
  journal={arXiv preprint arXiv:2106.02713},
  year={2021}
}

@article{schaipp2025surprising,
  title={The surprising agreement between convex optimization theory and learning-rate scheduling for large model training},
  author={Schaipp, Fabian and H{\"a}gele, Alexander and Taylor, Adrien and Simsekli, Umut and Bach, Francis},
  journal={arXiv preprint arXiv:2501.18965},
  year={2025}
}

@article{hagele2024scaling,
  title={Scaling laws and compute-optimal training beyond fixed training durations},
  author={H{\"a}gele, Alexander and Bakouch, Elie and Kosson, Atli and Allal, Loubna B and Von Werra, Leandro and Jaggi, Martin},
  journal={Advances in Neural Information Processing Systems},
  volume={37},
  pages={76232--76264},
  year={2024}
}

@article{shulgin2026deriving,
  title={Deriving hyperparameter scaling laws via modern optimization theory},
  author={Shulgin, Egor and von R{\"u}tte, Dimitri and Zhang, Tianyue H and Ajroldi, Niccol{\`o} and Sch{\"o}lkopf, Bernhard and Orvieto, Antonio},
  journal={arXiv preprint arXiv:2603.15958},
  year={2026}
}

@inproceedings{li2017stochastic,
  title={Stochastic modified equations and adaptive stochastic gradient algorithms},
  author={Li, Qianxiao and Tai, Cheng and others},
  booktitle={International Conference on Machine Learning},
  pages={2101--2110},
  year={2017},
  organization={PMLR}
}

@article{malladi2022sdes,
  title={On the SDEs and scaling rules for adaptive gradient algorithms},
  author={Malladi, Sadhika and Lyu, Kaifeng and Panigrahi, Abhishek and Arora, Sanjeev},
  journal={Advances in Neural Information Processing Systems},
  volume={35},
  pages={7697--7711},
  year={2022}
}

@article{compagnoni2024adaptive,
  title={Adaptive methods through the lens of SDEs: Theoretical insights on the role of noise},
  author={Compagnoni, Enea Monzio and Liu, Tianlin and Islamov, Rustem and Proske, Frank Norbert and Orvieto, Antonio and Lucchi, Aurelien},
  journal={arXiv preprint arXiv:2411.15958},
  year={2024}
}

@misc{compagnoni2026interaction,
      title={On the Interaction of Batch Noise, Adaptivity, and Compression, under $(L_0,L_1)$-Smoothness: An SDE Approach}, 
      author={Enea Monzio Compagnoni and Rustem Islamov and Frank Norbert Proske and Aurelien Lucchi and Antonio Orvieto and Eduard Gorbunov},
      year={2026},
      eprint={2506.00181},
      archivePrefix={arXiv},
      primaryClass={cs.LG},
      url={https://arxiv.org/abs/2506.00181}, 
}

@inproceedings{compagnoni2023sde,
  title={An sde for modeling sam: Theory and insights},
  author={Compagnoni, Enea Monzio and Biggio, Luca and Orvieto, Antonio and Proske, Frank Norbert and Kersting, Hans and Lucchi, Aurelien},
  booktitle={International Conference on Machine Learning},
  pages={25209--25253},
  year={2023},
  organization={PMLR}
}

@article{zhang2026beyond,
  title={Beyond a Single Explanation of the Adam--SGD Gap},
  author={Zhang, Chenxiang and Islamov, Rustem and Compagnoni, Enea Monzio and Pang, Jun and Lucchi, Aurelien and Orvieto, Antonio},
  journal={arXiv preprint arXiv:2606.14259},
  year={2026}
}

@article{defazio2023optimal,
  title={Optimal linear decay learning rate schedules and further refinements},
  author={Defazio, Aaron and Cutkosky, Ashok and Mehta, Harsh and Mishchenko, Konstantin},
  journal={arXiv preprint arXiv:2310.07831},
  year={2023}
}

@article{defazio2026schedulefree+,
  title={Schedulefree+: Scaling learning-rate-free \& schedule-free learning to large language models},
  author={Defazio, Aaron},
  journal={arXiv preprint arXiv:2605.19095},
  year={2026}
}

@article{defazio2024road,
  title={The road less scheduled},
  author={Defazio, Aaron and Yang, Xingyu and Mehta, Harsh and Mishchenko, Konstantin and Khaled, Ahmed and Cutkosky, Ashok},
  journal={Advances in Neural Information Processing Systems},
  volume={37},
  pages={9974--10007},
  year={2024}
}

@article{paquette20244,
  title={4+ 3 phases of compute-optimal neural scaling laws},
  author={Paquette, Elliot and Paquette, Courtney and Xiao, Lechao and Pennington, Jeffrey},
  journal={Advances in Neural Information Processing Systems},
  volume={37},
  pages={16459--16537},
  year={2024}
}

@misc{dieuleveut2020bridging,
     title={Bridging the Gap between Constant Step Size Stochastic Gradient Descent and Markov Chains}, 
      author={Aymeric Dieuleveut and Alain Durmus and Francis Bach},
      year={2018},
      eprint={1707.06386},
      archivePrefix={arXiv},
      primaryClass={stat.ML},
      url={https://arxiv.org/abs/1707.06386}, 
}

@inproceedings{
cai2026does,
title={Does {LLM} Pre-Training Typically Occur at the Edge of Stability?},
author={Yuhang Cai and Haofeng Huang and Haodong Wen and Deyi Liu and Yiyuan Ma and Kaifeng Lyu},
booktitle={Workshop on Scientific Methods for Understanding Deep Learning},
year={2026},
url={https://openreview.net/forum?id=QSb05IuPsy}
}

@inproceedings{
keskar2017on,
title={On Large-Batch Training for Deep Learning: Generalization Gap and Sharp Minima},
author={Nitish Shirish Keskar and Dheevatsa Mudigere and Jorge Nocedal and Mikhail Smelyanskiy and Ping Tak Peter Tang},
booktitle={International Conference on Learning Representations},
year={2017},
url={https://openreview.net/forum?id=H1oyRlYgg}
}

\clearpage
\appendix

\section{Additional Plots}
\label{app:additional_plots}
\begin{figure}[!htp]
    \centering
    \includegraphics[width=1.0\linewidth]{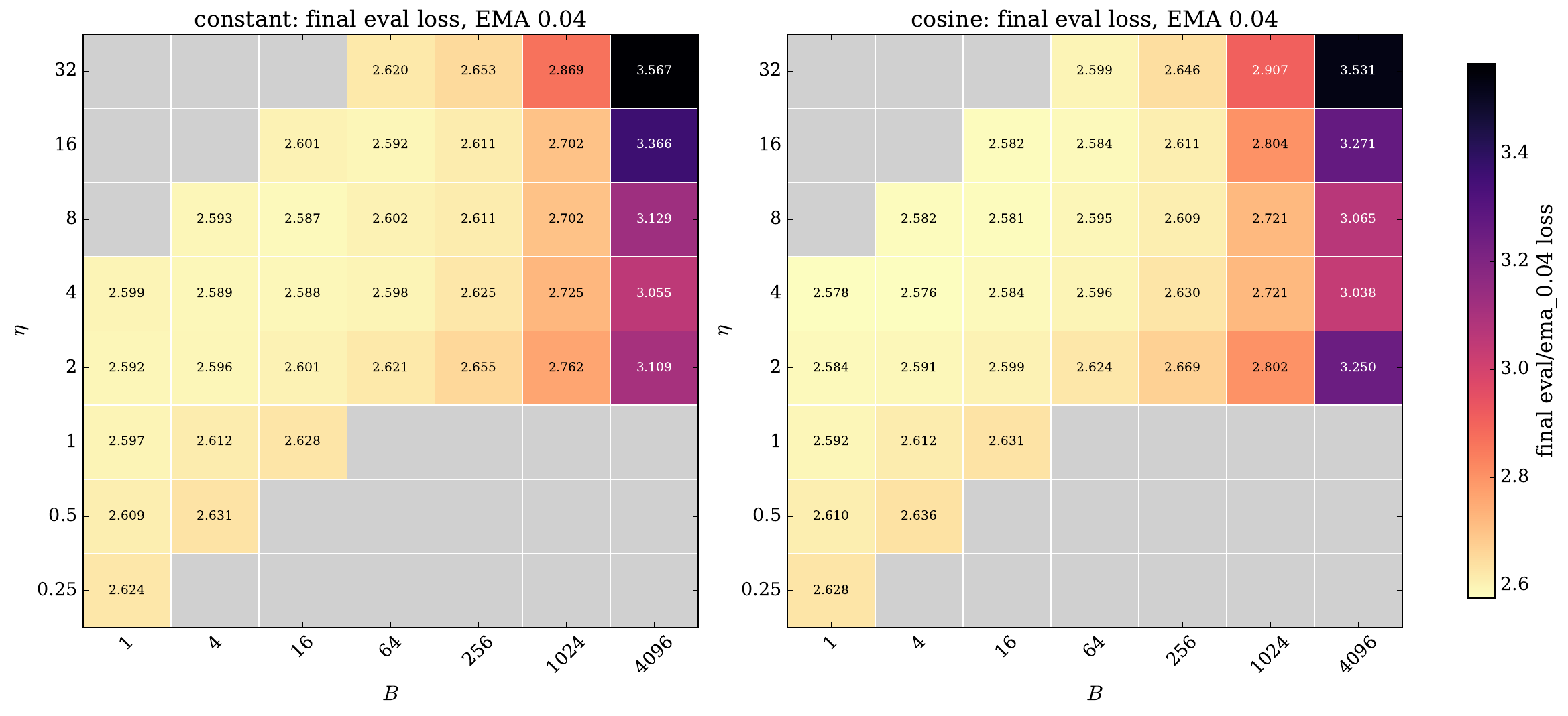}
    \caption{$(\eta, B)$ pretraining sweep for the constant with EMA and cosine trained models. }
    \label{fig:eta_B_pretraining_sweep}
\end{figure}

\begin{figure}
    \centering
\includegraphics[width=1.0\linewidth]{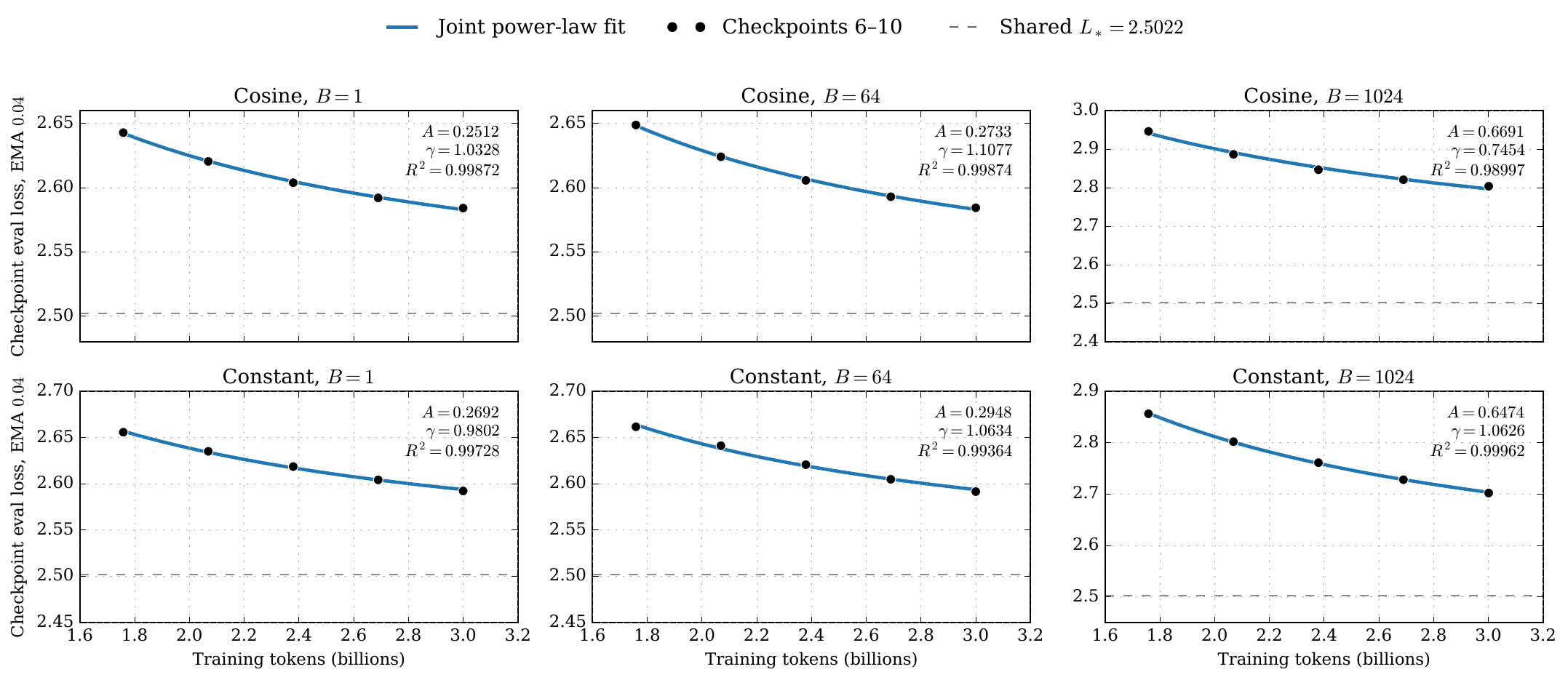}
    \caption{Joint power law fit for the loss as a function of the tokens of the form $L = L_\star + A T^{-\gamma}$.}\label{fig:joint_power_law_fit}
\end{figure}

\begin{figure}[!htp]
    \centering
    \includegraphics[width=1.0\linewidth]{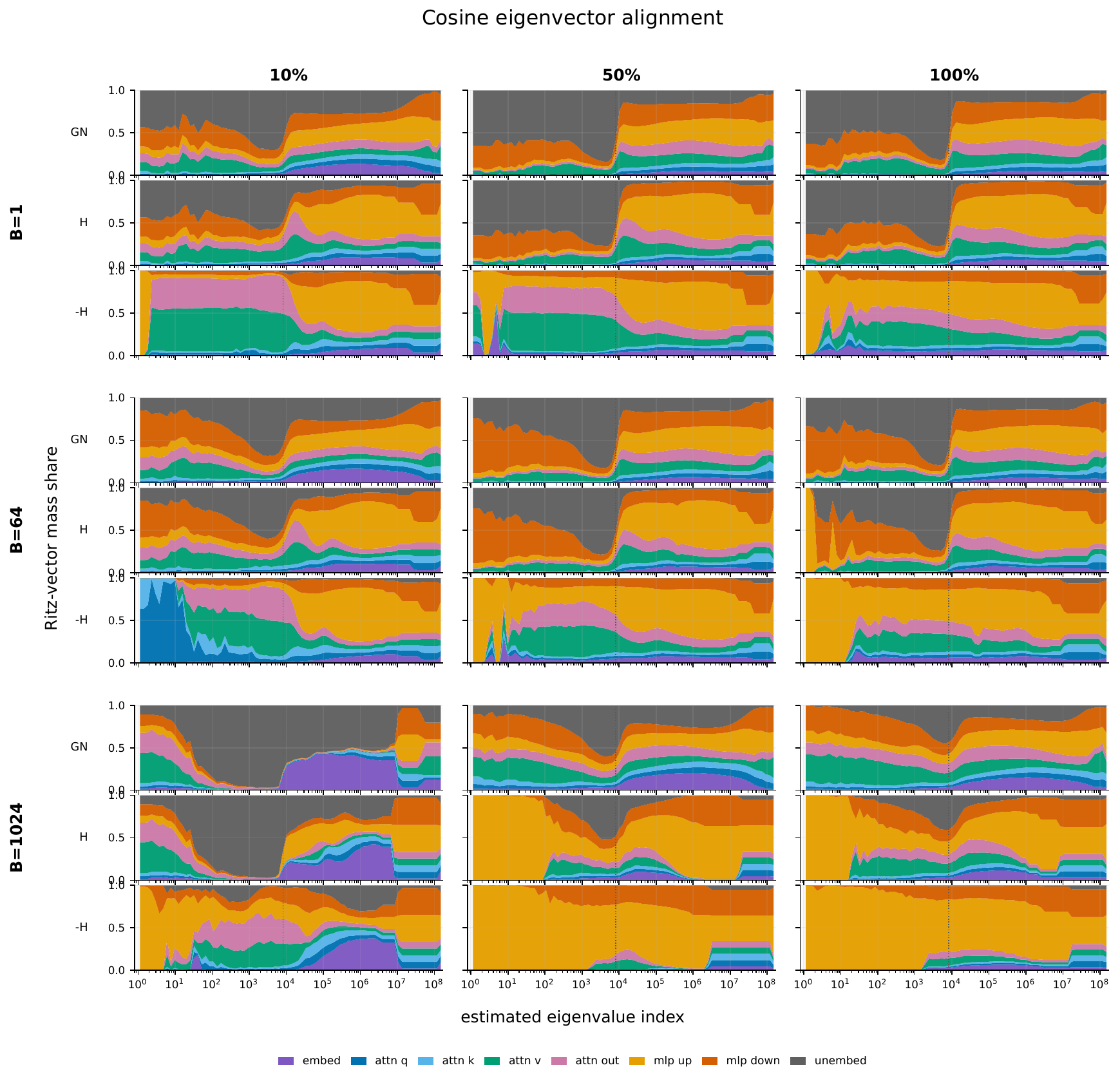}
    \caption{A full sweep of the eigenvector distribution at the end of training for cosine decay.}
    \label{fig:full_evec_cosine}
\end{figure}

\begin{figure}[!htp]
	\includegraphics[width=\textwidth]{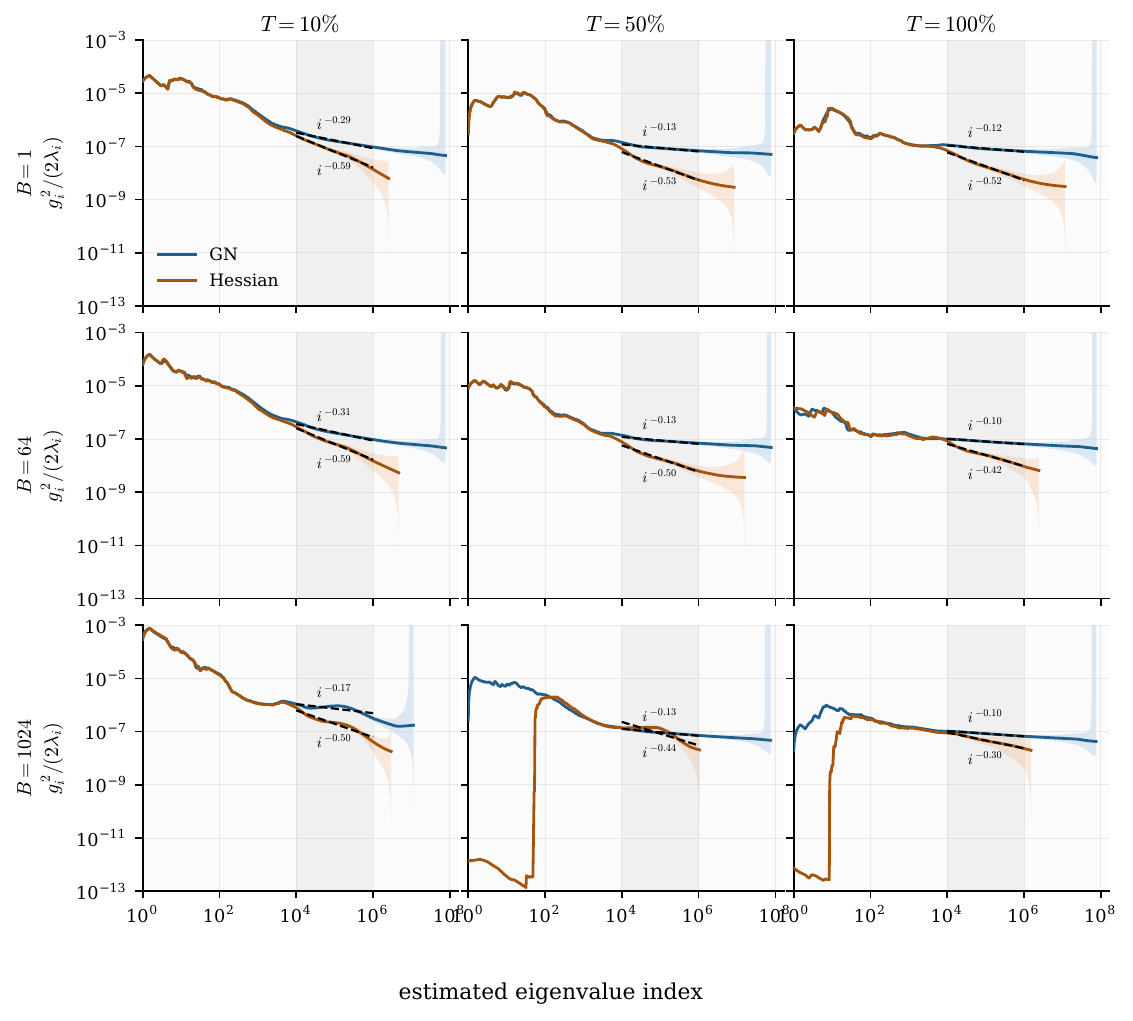}	
	\caption{An extended version of \Cref{fig:source_cosine} with additional checkpoints. This is for the set of runs using cosine decay after warmup. See \Cref{fig:source_constant_full} for the constant schedule runs.}
	\label{fig:source_cosine_full}
\end{figure}
\clearpage

\section{Constant Learning Rate with EMA}
\label[appendix]{app:constant_learning_rate}
All of the plots in the main paper were run with 10\% warmup followed by cosine decay to 10\% of the maximal learning rate. We also repeated the experiments with a constant learning rate schedule, which remains fixed after warmup.

\begin{figure}[!htp]
    \centering
    \includegraphics[width=1.0\linewidth]{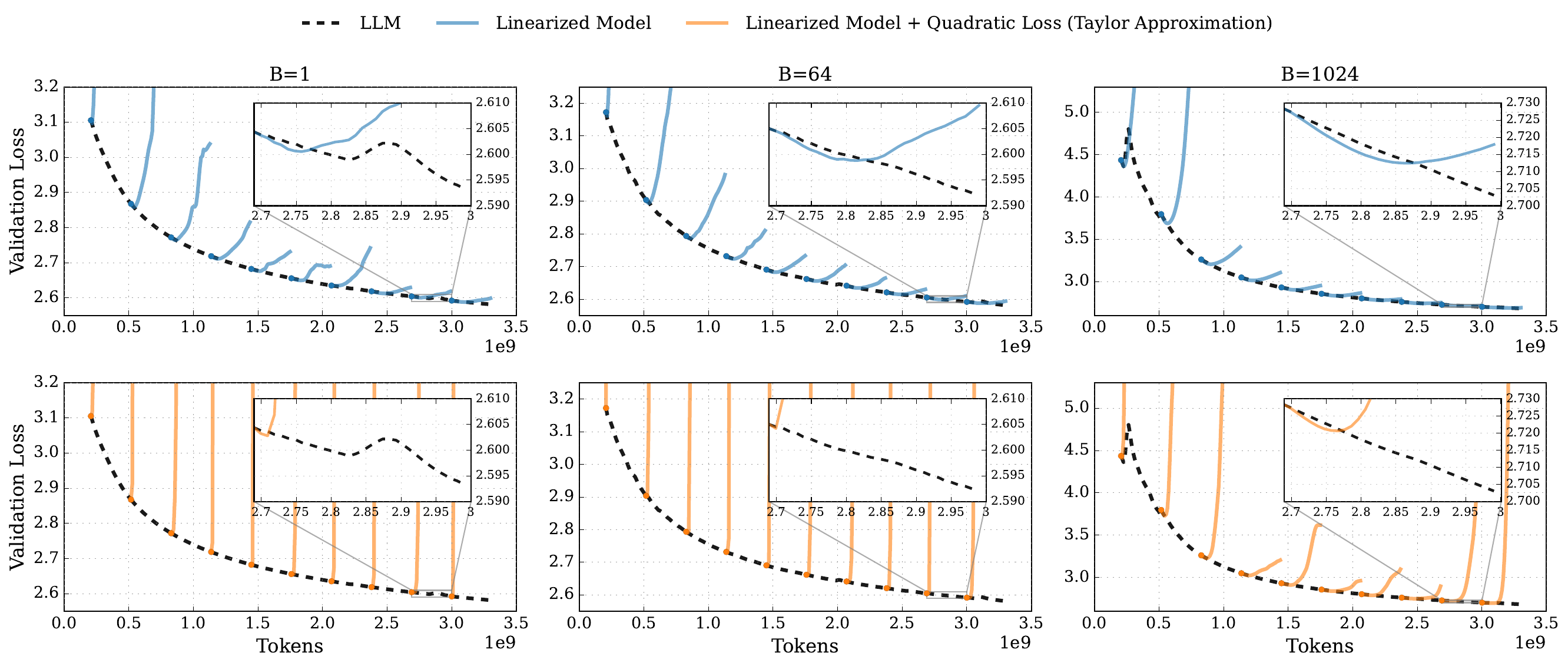}
    \caption{Validation loss curves for 150M models (black) trained with constant learning with EMA for $1 \times$ Chinchilla tokens, across batch sizes, at optimal learning rates, under the same setup as Figure~\ref{fig:cosine_linearization_loss_plots}. The quadratic expansions are much more unstable, and we leave a study of this to future work. We provide further details in Section~\ref{sec:quadratics_are_predictive}.}
    \label{fig:constant_linearization_loss_plots}
\end{figure}

\begin{figure}[!htp]
    \includegraphics[width=\textwidth]{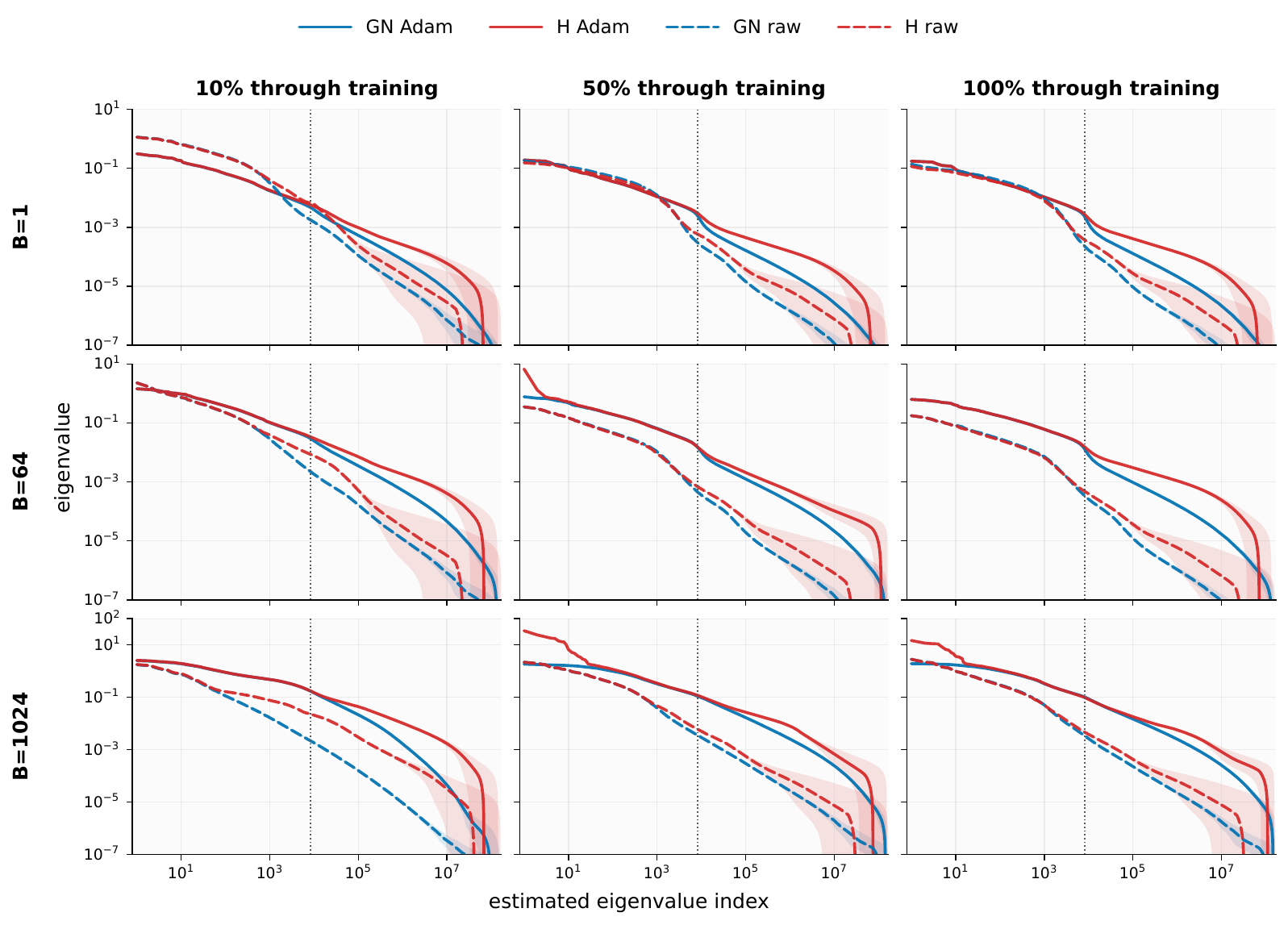}
    \caption{The spectra using a constant learning rate schedule after warmup, rather than cosine decay. This is otherwise equivalent to \Cref{fig:spectrum-summary}.}
       \label{fig:constant_spectrum}
\end{figure}

\begin{figure}[!htp]
    \centering
    \includegraphics[width=1.0\linewidth]{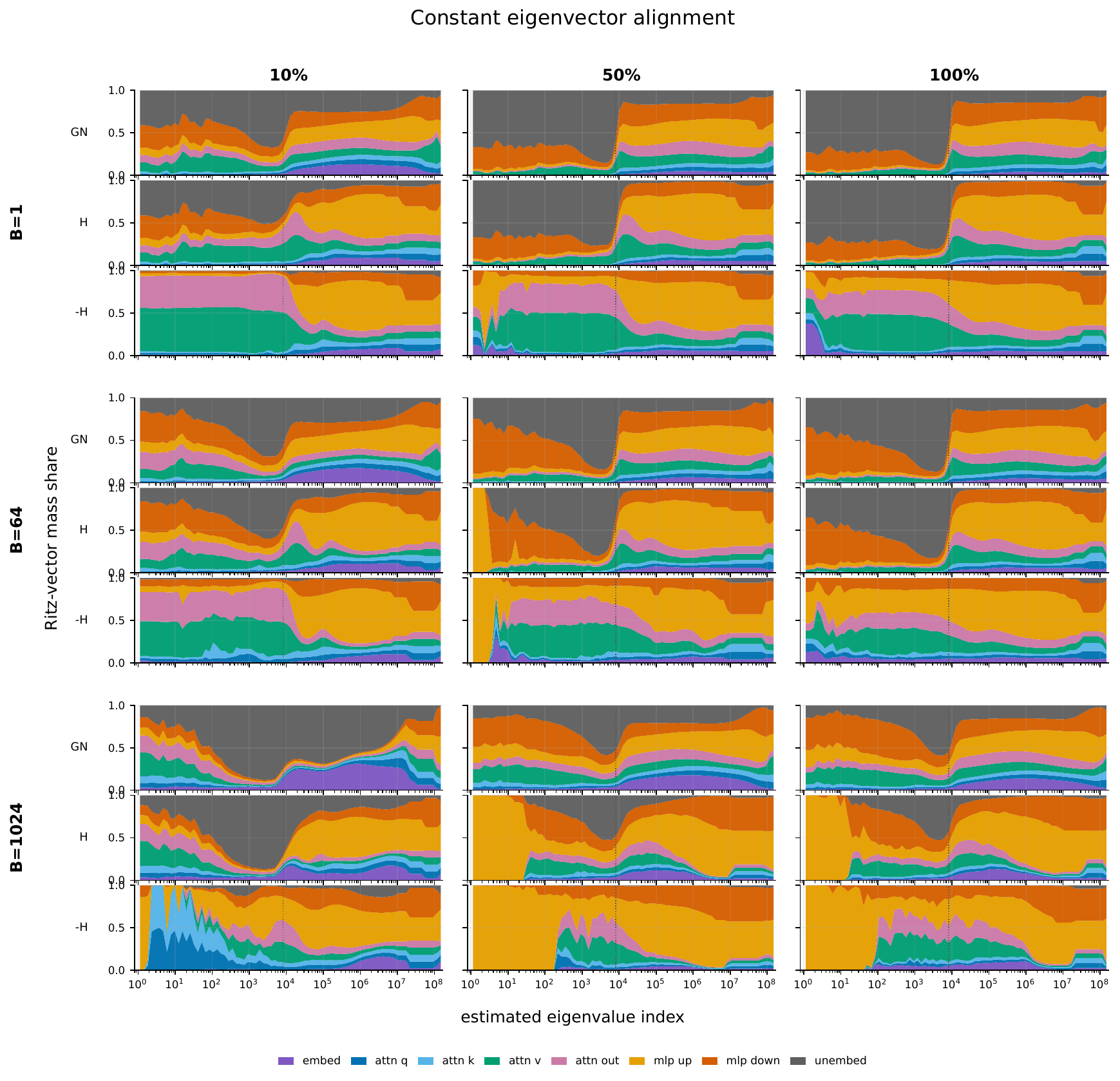}
    \caption{A full sweep of the eigenvector distribution at the end of training for the constant learning rate schedule. This is otherwise equivalent to \Cref{fig:evecs_cosine,fig:full_evec_cosine}.}
    \label{fig:full_evec_constant}
\end{figure}

\begin{figure}[!htp]
	\includegraphics[width=\textwidth]{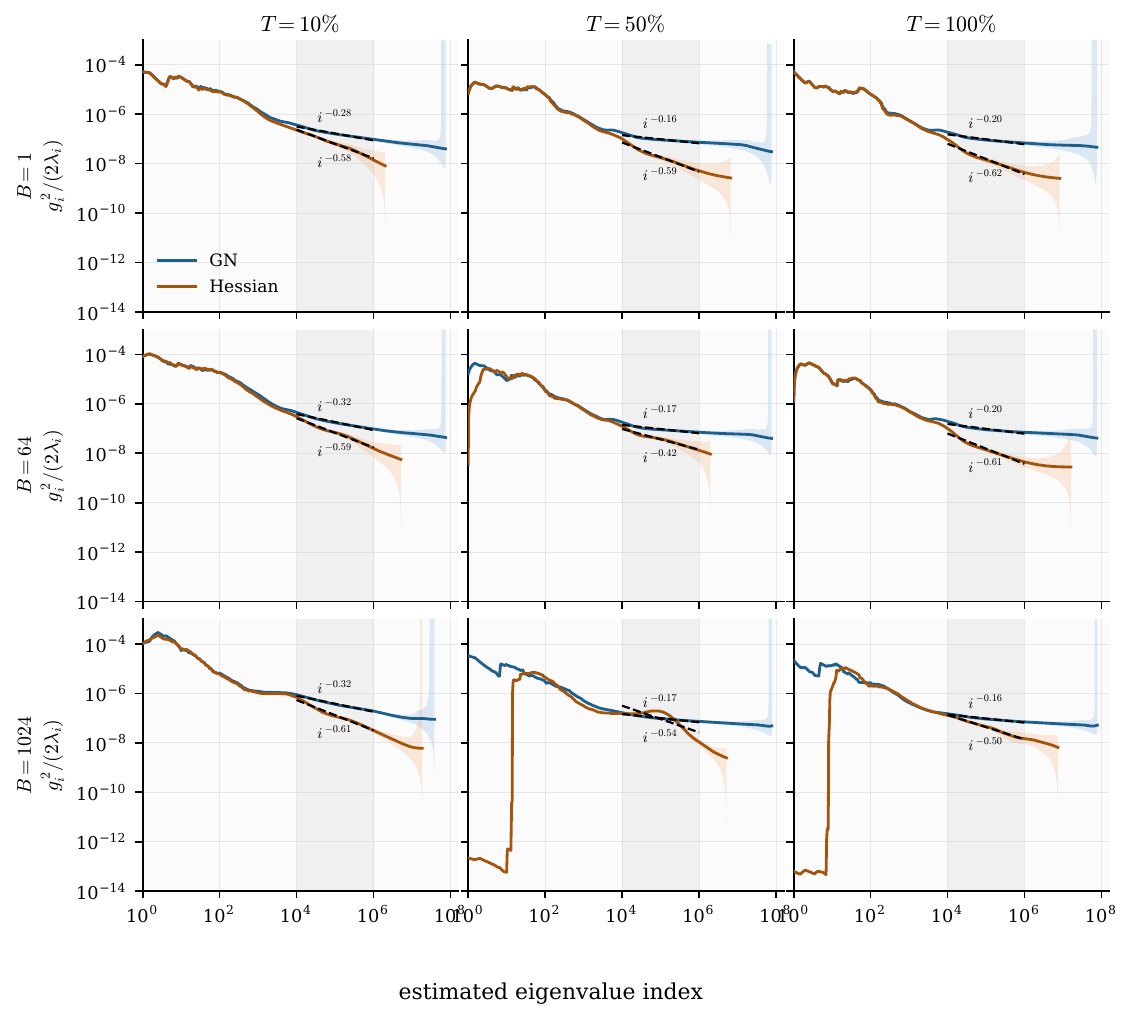}	
	\caption{The source condition using a constant learning rate schedule after warmup, rather than cosine decay. This is otherwise equivalent to \Cref{fig:source_cosine,fig:source_cosine_full}.}
	\label{fig:source_constant_full}
\end{figure}

\begin{figure}[!htp]
    \centering
    \includegraphics[width=1.0\linewidth]{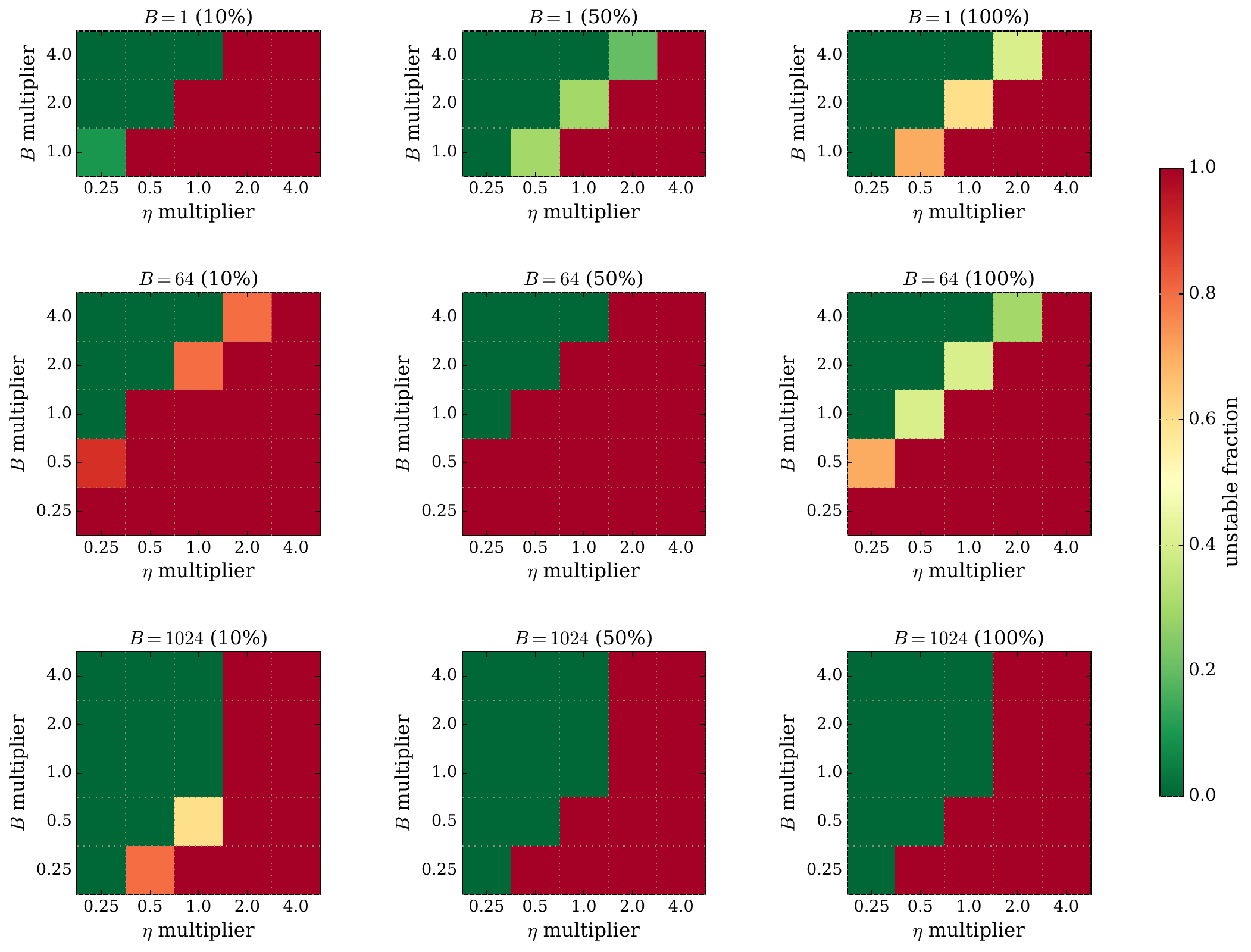}
    \caption{Heatmap showing instability for models trained with constant learning rate with EMA under the same setup as Figure~\ref{fig:cosine_linearization_loss_plots}. As in \Cref{fig:cosine_linearization_loss_plots}, the smaller batch sizes operate near the stochastic edge of stability. However, because the stable region in the $B=1024$ plots is diagonal then vertical, it operates at both the deterministic and stochastic edge of stability. This is the case whenever $(\eta,B)$ is stable but both $(2\eta,2B)$ and $(\eta,B/2)$ are unstable. This is distinct from our cosine schedule experiments for which $B=1024$ operated entirely at the deterministic edge of stability.}
    \label{fig:eos_constant_grid}
\end{figure}

\begin{figure}
    \centering
    \includegraphics[width=\linewidth]{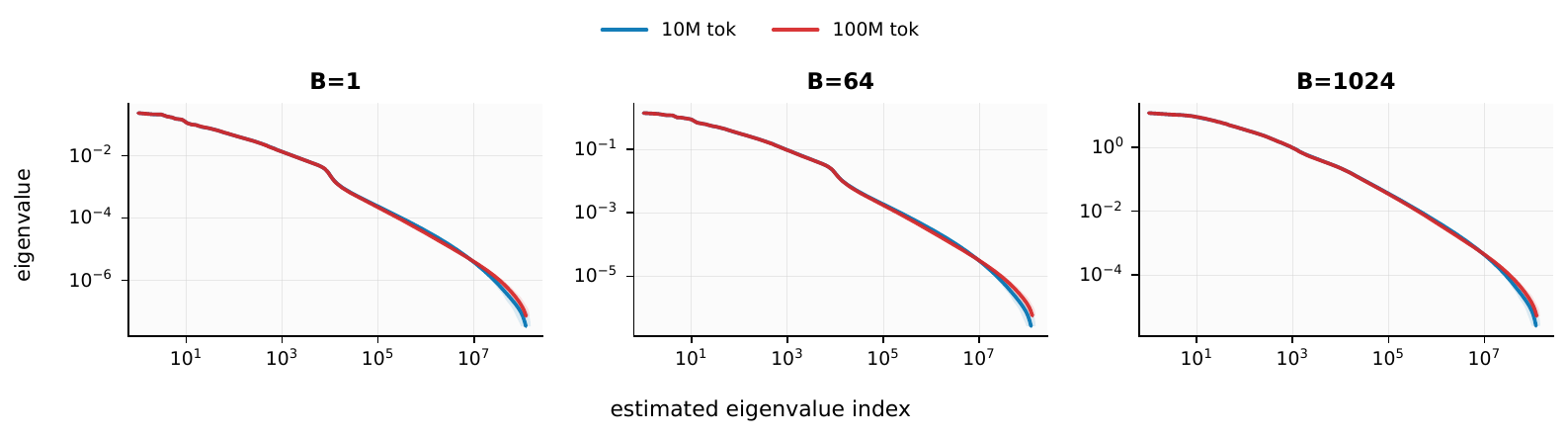}
    \caption{Comparison of the preconditioned Gauss Newton spectrum at $T=100\%$ computed over 10M and 100M tokens, both at sequence length $1024$. Each plot shows a different pretraining batch size, which is distinct from the number of samples used to estimate the Hessian. The spectra largely agree, even deep in the tail, so our remaining experiments estimate the Hessian over 10M tokens.}
    \label{fig:100m_comparison}
\end{figure}

\begin{figure}
    \centering
    \includegraphics[width=\linewidth]{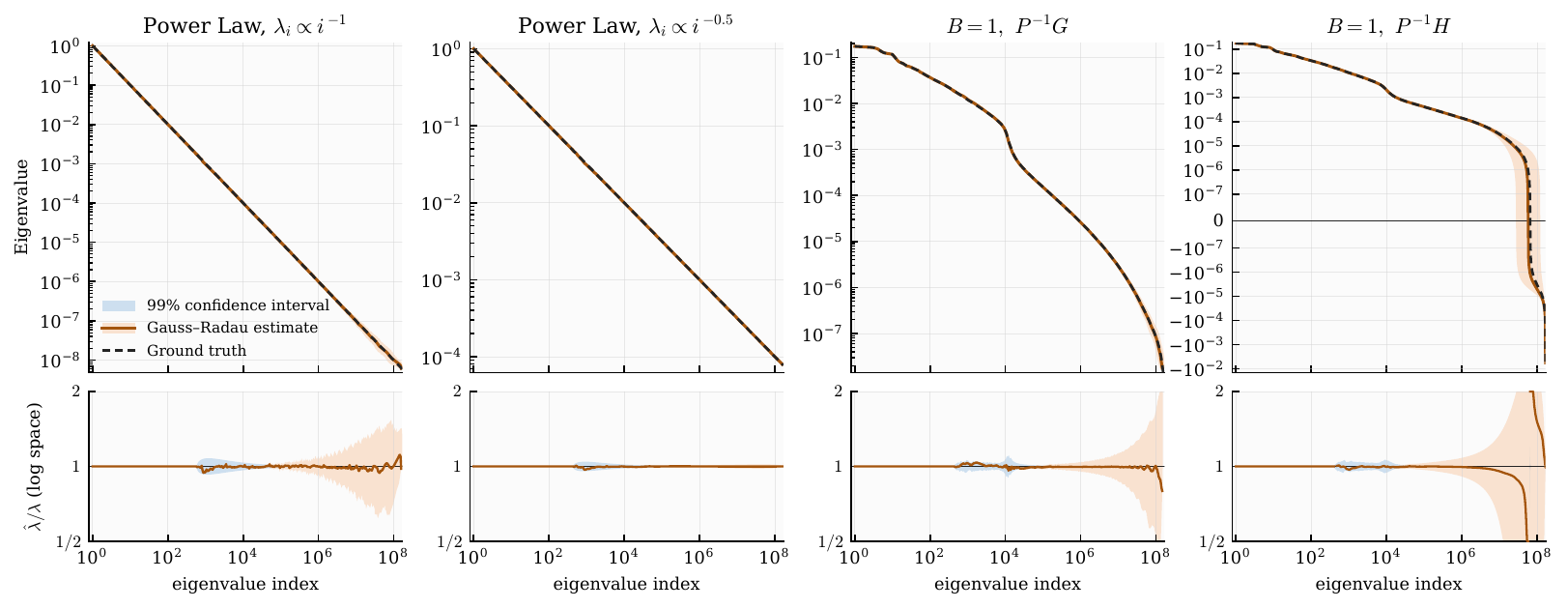}
    \caption{Comparing the ground truth (dashed black) with the SLQ estimate (orange) for various spectra. For the ground truth, we tested two power law spectra $\lambda_i \propto i^{-\alpha}$ along with the estimated spectra for $P^{-1} G$ and $P^{-1} H$ to check consistency with an additional SLQ probe. The light orange bands show the Gauss-Radau error bands from \Cref{alg:gauss_radau_quadrature} while the light blue bands show a 99\% confidence interval for the error that comes from sampling a single random probe. The bottom row shows that the relative error $\log(\hat \lambda_i/\lambda_i)$ remains small deep in the spectrum.}
    \label{fig:spectrum_bootstrap}
\end{figure}

\clearpage

\end{document}